\documentclass[sigconf]{acmart}

\usepackage{subfigure}
\usepackage[linesnumbered,ruled,noend]{algorithm2e} 
\usepackage{multirow}

\AtBeginDocument{%
  \providecommand\BibTeX{{%
    \normalfont B\kern-0.5em{\scshape i\kern-0.25em b}\kern-0.8em\TeX}}}

\copyrightyear{2023}
\acmYear{2023}
\setcopyright{acmlicensed}
\acmConference[KDD '23] {Proceedings of the 29th ACM SIGKDD Conference on Knowledge Discovery and Data Mining}{August 6--10, 2023}{Long Beach, CA, USA.}
\acmBooktitle{Proceedings of the 29th ACM SIGKDD Conference on Knowledge Discovery and Data Mining (KDD '23), August 6--10, 2023, Long Beach, CA, USA}
\acmPrice{15.00}
\acmISBN{979-8-4007-0103-0/23/08}
\acmDOI{10.1145/3580305.3599469}

\begin{document}

\title{Precursor-of-Anomaly Detection for Irregular Time Series}

\author{Sheo Yon Jhin}
\affiliation{%
  \institution{Yonsei University}
  \city{Seoul}
  \country{South Korea}
}\email{sheoyonj@yonsei.ac.kr}

\author{Jaehoon Lee}
\authornote{This work was done when he was at Yonsei university.}
\affiliation{%
  \institution{LG AI Research \& Yonsei University}
  \city{Seoul}
  \country{South Korea}}
\email{jaehoon.lee@lgresearch.ai}

\author{Noseong Park}
\affiliation{%
  \institution{Yonsei University}
  \city{Seoul}
  \country{South Korea}
}
\email{noseong@yonsei.ac.kr}

\renewcommand{\shortauthors}{Sheo Yon Jhin, Jaehoon Lee, \& Noseong Park}
\begin{abstract}

Anomaly detection is an important field that aims to identify unexpected patterns or data points, and it is closely related to many real-world problems, particularly to applications in finance, manufacturing, cyber security, and so on. While anomaly detection has been studied extensively in various fields, detecting future anomalies before they occur remains an unexplored territory. In this paper, we present a novel type of anomaly detection, called \emph{\textbf{P}recursor-of-\textbf{A}nomaly} (PoA) detection. Unlike conventional anomaly detection, which focuses on determining whether a given time series observation is an anomaly or not, PoA detection aims to detect future anomalies before they happen. To solve both problems at the same time, we present a neural controlled differential equation-based neural network and its multi-task learning algorithm. We conduct experiments using 17 baselines and 3 datasets, including regular and irregular time series, and demonstrate that our presented method outperforms the baselines in almost all cases. Our ablation studies also indicate that the multitasking training method significantly enhances the overall performance for both anomaly and PoA detection.
\end{abstract}




\received{20 February 2007}
\received[revised]{12 March 2009}
\received[accepted]{5 June 2009}

\maketitle

\section{Introduction}
\begin{figure}
    \centering
    \subfigure[Anomaly detection]{\includegraphics[width=0.35\columnwidth]{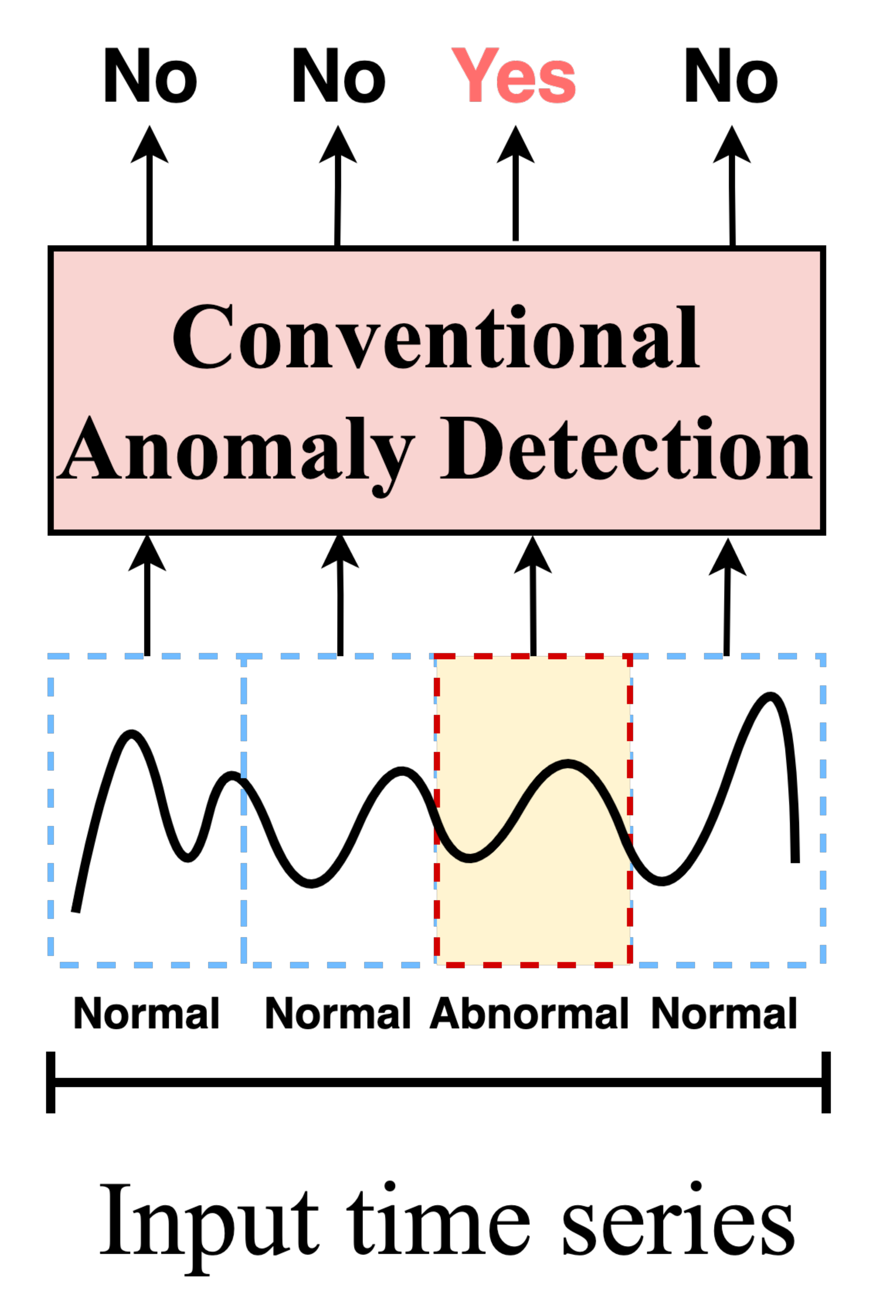}} \hspace{2em}
    \subfigure[PoA detection]{\includegraphics[width=0.35\columnwidth]{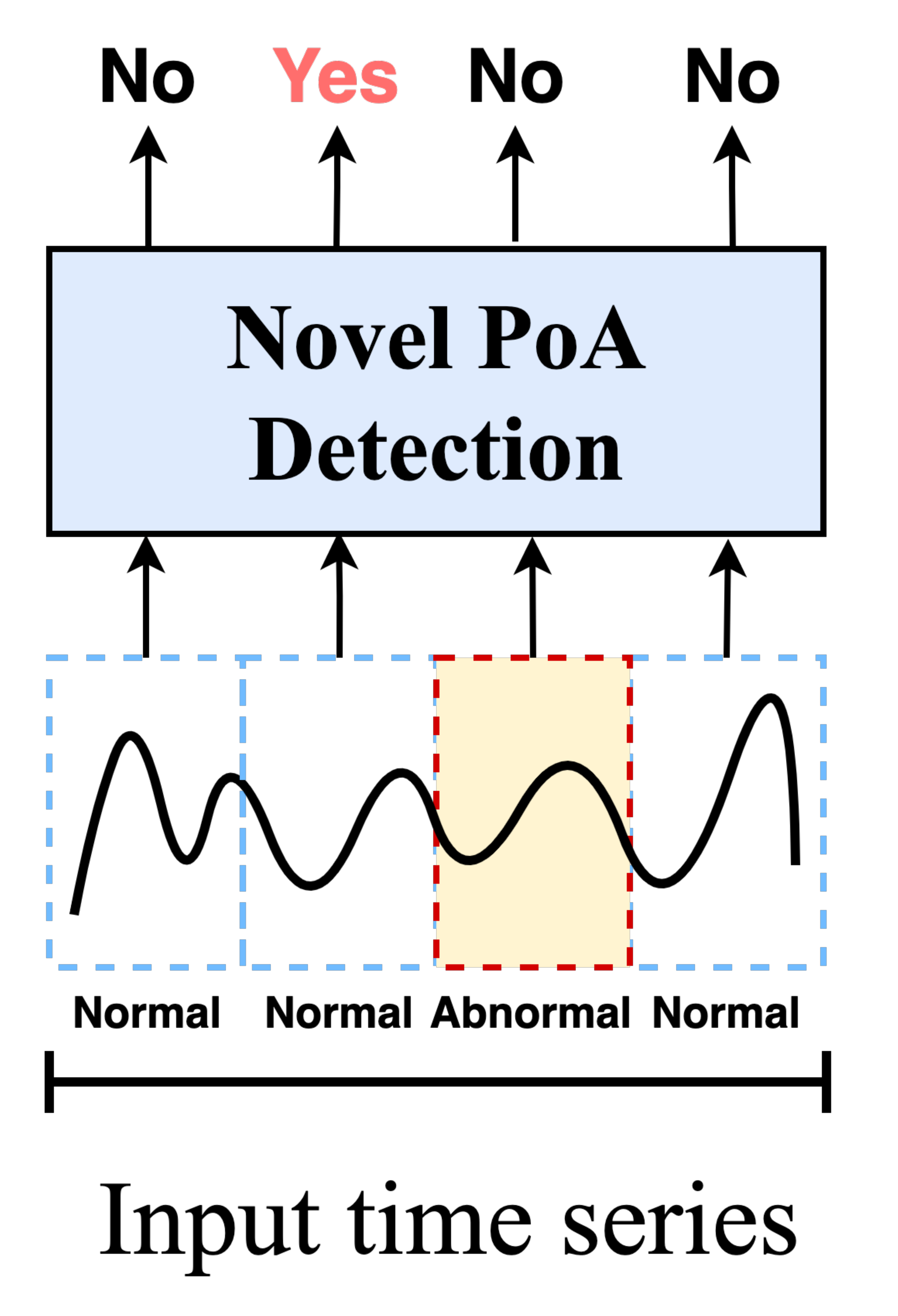}}
    \caption{Comparison between the conventional anomaly detection and our proposed the precursor-of-anomaly (PoA) detection. In PoA, we predict whether the next window will contain any abnormal observation before it happens, which is much more challenging than the anomaly detection.}
    \label{fig:comparison1} 
\end{figure}
Anomaly detection is an important and difficult task that has been extensively studied for many real-world applications~\citep{eskin2002geometric,zhang2013medmon,ten2011anomaly,goldstein2016comparative}. The goal of anomaly detection is to find unexpected or unusual data points and/or trends that may indicate errors, frauds, or other abnormal situations requiring further investigations.

\paragraph{\textbf{Novel task definition:}} Among many such studies, one of the most popular setting is \emph{multivariate time series anomaly detection} since many real-world applications deal with time series, ranging from natural sciences, finance, cyber security, and so on. Irregular time series anomaly detection~\citep{wu2019lstm,liu2020deep,chandola2010anomaly} is of utmost importance since time series data is frequently irregular. Many existing time series anomaly detection designed for regular time series show sub-optimal outcomes when being applied to irregular time series. Irregular time series typically have complicated structures with uneven inter-arrival times.

In addition, it is also important to forecast whether there will be an anomaly in the future given a time series input, which we call \emph{\textbf{P}recursor-of-\textbf{A}nomaly} (PoA) detection. The precursor-of-anomaly detection refers to the process of identifying current patterns/signs that may indicate upcoming abnormal events. The goal is to detect these precursors before actual anomalies occur in order to take preventive actions (cf. Fig.~\ref{fig:comparison1}). The precursor-of-anomaly detection can be applied to various fields such as finance, medical care, and geophysics. For example, in finance, the precursor-of-anomaly detection can be used to identify unusual patterns in stock prices that may indicate a potential market collapse, and in geophysics, the precursor-of-anomaly detection can detect unusual geometric activities that can indicate earthquakes. Despite of its practical usefulness, the precursor-of-anomaly detection has been overlooked for a while due to its challenging nature except for a handful of work to detect earthquakes based on manual statistical analyses~\citep{bhardwaj2017review,saraf2009advances,ghosh2009anomalous}.

To this end, we mainly focus on the following two tasks for multivariate and irregular time series: i) detecting whether the input time series sample has anomaly or not, i.e., detecting an anomaly, and ii) detecting whether anomaly will happen after the input time series sample, i.e., detecting the precursor-of-anomaly. In particular, we are the first detecting the precursor-of-anomaly for general domains. Moreover, we solve the two related problems at the same time with a single framework.

\paragraph{\textbf{Novel method design:}} Our work is distinctive from existing work not only in its task definition but also in its deep learning-based method design. Recent anomaly detection methods can be divided into three main categories: i) clustering-based ii) density estimation-based iii) reconstruction-based methods. Clustering-based methods use historical data to identify and measure the distance between normal and abnormal clusters but can struggle to detect anomalies in complex data where distinct abnormal patterns are unknown~\citep{agrawal2015survey}. Density estimation-based methods estimate a density function to detect anomalies. These methods assume that abnormal samples lie in a low density region. However, there is a limitation in data where the difference between low-probability normal and abnormal samples is not clear~\citep{zong2018deep}. On the other hand, reconstruction-based methods aim to identify abnormal patterns by learning low-dimensional representations of data and detecting anomalies based on reconstruction errors. However, these methods may have limitations in modeling dependencies and temporal dependencies in time series, making them unsuitable for detecting certain types of anomalies~\citep{an2015variational,zenati2018efficient}. To detect anomalies that may not be apparent in raw time-series, there are several advanced methods~\citep{moghaddass2019anomaly,knorn2008adaptive}.

In this paper, we propose i) a unified framework of detecting the anomaly and precursor-of-anomaly based on neural controlled differential equations (NCDEs) and ii) its multi-task learning and knowledge distillation-based training algorithm with self-supervision. NCDEs are recently proposed differential equation-based \emph{continuous-time recurrent neural networks} (RNNs), which have been widely used for irregular time series forecasting and classification. For the first time, we adopt NCDEs to time series (precursor of) anomaly detection. Our method, called (\underline{\textbf{P}}recursor of) \underline{\textbf{A}}nomaly \underline{\textbf{D}}etection (PAD), is able to provide both the anomaly and the precursor-of-anomaly detection capabilities.

As shown in Fig.~\ref{fig:overall} there are two co-evolving NCDEs: i) the anomaly NCDE layer for the anomaly detection, ii) and the PoA NCDE layer for the precursor-of-anomaly detection. They are trained as a single unified framework under the multi-task training and knowledge distillation regime. Since our method is trained in a self-supervised manner, we resample the normal training dataset to create artificial anomalies (see Section ~\ref{sec:data_augmentation}). To train the PoA NCDE layer, we apply a knowledge distillation method where the anomaly NCDE layer becomes as a teacher and the PoA NCDE becomes a student. By conducting a training by knowledge distillation, the PoA NCDE can inherit the knowledge of the anomaly NCDE. In other words, the PoA NCDE layer, which reads a temporal window up to time $i$, mimics the inference of the anomaly NCDE which reads a window up to time $i+1$ and decide whether it is an anomaly, which is a possible PoA detection approach since the PoA NCDE layer sees only past information. The two different NCDEs co-evolve while interacting with each other. At the end, our proposed method is trained for all the two tasks. They also partially share trainable parameters, which is a popular the multi-task design.

We performed experiments with 17 baselines and 3 datasets. In addition to the experiments with full observations, we also test after dropping random observations in order to create challenging detection environments. In almost all cases, our proposed method, called \emph{PAD}, shows the best detection ability. Our ablation studies prove that the multi-task learning method enhances our method's capability. We highlight the following contributions in this work: 
\begin{enumerate}
    \item We are the first solving the anomaly and the PoA detection at the same time.
    \item We propose PAD which is an NCDE-based unified framework for the anomaly and the PoA detection.
    \item To this end, we also design a multi-task and knowledge distillation learning method.
    \item In addition, the above learning method is conducted in a self-supervised manner. Therefore, we also design an augmentation method to create artificial anomalies for our self-supervised training.
    \item We perform a comprehensive set of experiments, including various irregular time series experiments. Our PAD marks the best accuracy in almost all cases.
    \item Our code is available at this link \footnote{https://github.com/sheoyon-jhin/PAD}, and we refer readers to Appendix for the information on reproducibility.
    \end{enumerate}

\section{Related Work}
\subsection{Anomaly Detection in Time Series}

Anomaly detection in time series data has been a popular research area in the fields of statistics, machine learning and data mining. Over the years, various techniques have been proposed to identify anomaly patterns in time series data, including machine learning algorithms, and deep learning models.

Machine learning algorithms and deep learning-based anomaly detection methods in time series can be categorized into 4 methods including i) classical methods, ii) clustering-based methods, iii) density-estimation methods, iv) reconstruction-based methods. 

Classical methods find unusual samples using traditional machine learning methods such as OCSVM and Isolation Forest~\citep{tax2004support,liu2008isolation}. 
Clustering-based methods are a type of unsupervised machine learning method for detecting anomalies in time-series data such as Deep SVDD, ITAD, and THOC~\citep{ruff2019deep,shin2020itad,shen2020timeseries}. This method splits the data into multiple clusters based on the similarity between the normal data points and then identifies abnormal data points that do not belong to any clusters. However, clustering-based methods are not suitable for complex data to train. Density-estimation-based methods~\citep{breunig2000lof,yairi2017data} are a type of unsupervised machine learning method for detecting anomalies in time series data such as LOF. In density-estimation based anomaly detection, the time-series data is transformed into a feature space, and the probability density function of the normal data points is estimated using techniques such as kernel density estimation or gaussian mixture models. The data points are then ranked based on their density, and anomalies are identified as data points that have low density compared to the normal data points. However, this method may not perform well in cases where the data contains significant non-stationary patterns or where the anomalies are not well represented by the estimated density function. Additionally, this method can also be computationally expensive, as it requires estimating the density function for the entire data. Reconstruction-based methods are~\citep{park2018multimodal,su2019robust, audibert2020usad, li2021multivariate} a representative methodology for detecting anomalies in time series data such as USAD~\citep{audibert2020usad}. In this method, a deep learning model reconstructs normal time-series data and then identifies abnormal time-series data with high reconstruction errors. Also, reconstruction-based methods can be effective in detecting anomalies in time-series data, especially when the anomalies are not well separated from the normal data points and the data contains complex patterns. The reconstruction-based methods are also robust to missing values and noisy data, as the reconstructed data can be used to fill in missing values and reduce the impact of noise. However, this method can be sensitive to the choice of reconstruction model, and may not perform well in cases where the anomalies are not well represented by the reconstruction model. Additionally, this method may be computationally expensive, as it requires training a complex model on the data.

In this paper, we mainly focus on self-supervised learning-based anomaly detection. Anomaly detection methods based on self-supervised learning have been extensively studied~\citep{cutpaste2021,puzzleae,ijcai2022p394,https://doi.org/10.48550/arxiv.2007.08176}. Unlike the unsupervised-learning method, self-supervised learning learns using negative samples generated by applying data augmentation to the training data~\citep{ijcai2022p394,https://doi.org/10.48550/arxiv.2007.08176}. Data augmentation can overcome the limitations of existing training data. In this paper, we train our method, PAD, on time series dataset generated by data augmentation. 


\begin{figure}
    \centering
    \includegraphics[width=0.4\columnwidth]{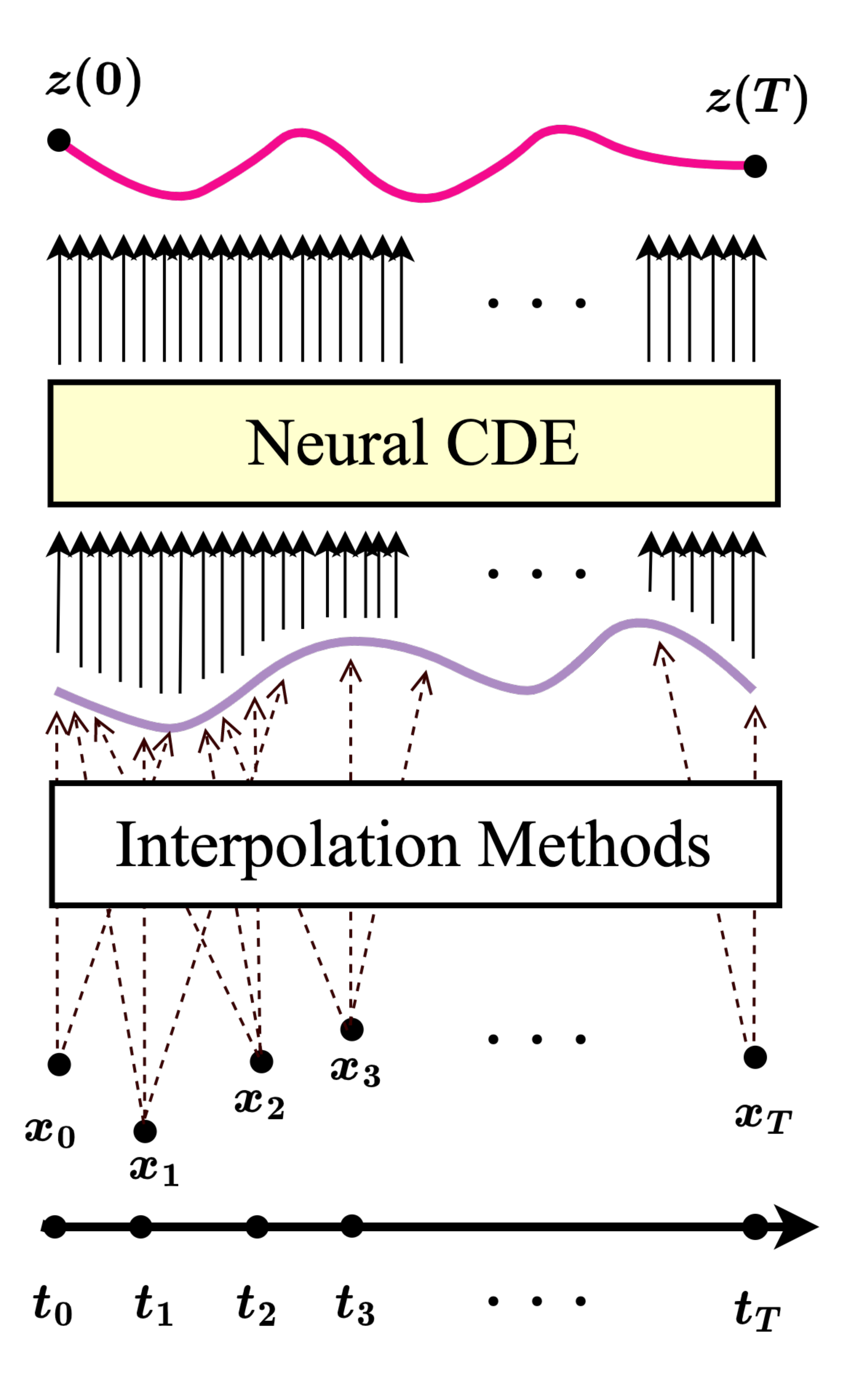}
    \caption{The architecture of NCDEs.}
    \label{fig:ncde}
\end{figure}

\subsection{Neural Controlled Differential Equations}
Recently, differential equation-based deep learning models have been actively researched. Neural controlled differential equations (NCDEs) are a type of machine learning method for modeling and forecasting time-series data. NCDE combines the benefits of neural networks and differential equations to create a flexible and powerful model for time-series data.
Unlike traditional time series models (e.g., RNNs), differential equation-based neural networks estimate the continuous dynamics of hidden vectors $\mathbf{z(t)}$ (i.e., $\frac{d\mathbf{z(t)}}{dt}$). Neural ordinary differential equations (NODEs) use the following equations to model a hidden dynamics~\citep{NIPS2018_7892}:
\begin{align}
    \mathbf{z}(T) = \mathbf{z}(0) + \int_{0}^{T}f(\mathbf{z}(t),t;\mathbf{\theta}_f)dt.
\end{align}

In contrast to NODEs, neural controlled differential equations (NCDEs) utilize the riemann–stieltjes integral~\citep{NEURIPS2020_4a5876b4}. Let $\{\mathbf{x}_i\}_{i=0}^N$ be time series observations whose length is $N$ and $\{t_i\}_{i=0}^N$ be its observation time-points. NCDEs can be formulated as the following equations:
\begin{align}
\mathbf{z}(T) &= \mathbf{z}(0) + \int_{0}^{T} f(\mathbf{z}(t);\mathbf{\theta}_f) dX(t),\\
 &=\mathbf{z}(0) + \int_{0}^{T} f(\mathbf{z}(t);\mathbf{\theta}_f) \frac{dX(t)}{dt} dt,\label{eq:ncde}
\end{align} where $X(t)$ is a continuous time series path  interpolated from $\{(\mathbf{x}_i, \mathbf{t}_i)\}_{i=0}^N$. With interpolation methods, we can obtain a continuous time series path from discrete observations $\{(\mathbf{x}_i, \mathbf{t}_i)\}_{i=0}^N$. Typically, natural cubic spline~\citep{mckinley1998cubic} is used as the interpolation method --- note that the adaptive step size solver works properly if the path $X(t)$ is twice continuously differentiable~\citep{NEURIPS2020_4a5876b4}.

The main difference between NODEs and NCDEs is the existence of the continuous time series path $X(t)$. NCDEs process a hidden vector $\mathbf{z}(t)$ along the interpolated path $X(t)$. (cf. Fig.~\ref{fig:ncde}) In this regard, NCDEs can be regarded as a continuous analogue to RNNs. Because there exists $X(t)$, NCDEs can learn what NODEs cannot (See Theorem C.1 in~\citep{NEURIPS2020_4a5876b4}). Additionally, NCDEs have the ability to incorporate domain-specific knowledge into the model, by using expert features as inputs to the neural networks. This can improve the performance and robustness of the model, as well as provide a deeper understanding of the relationships between the variables in the data. Therefore, NCDEs enable us to model more complex patterns in continuous time series. 

\subsection{Knowledge Distillation} 
Knowledge distillation~\citep{hinton2015distilling} is a technique used to transfer knowledge from large, complex models to smaller, simpler models. In the context of time series data, knowledge distillation can be used to transfer knowledge from complex deep learning models to smaller, more efficient models.
The knowledge extraction process involves training complex models such as deep neural networks on large datasets of time series data. The complex model is then used to generate predictions for the same data set. These predictions are then used to train simpler models such as linear regression models or decision trees. In this paper, we use these predictions in precursor-of-anomaly detection.
During the training process, simpler models are optimized to minimize the difference between their predictions and those of complex models. This allows simple models to learn patterns and relationships in time series data captured by complex models.
Furthermore, knowledge distillation can also help improve the interpretability of the model. The simpler model can be easier to understand and analyze, and the knowledge transferred from the complex model can provide insights into the important features and relationships in the time-series data.
The advantages of distilling knowledge from time series include faster and more efficient model training, reduced memory requirements, and improved interpretability of the model~\citep{ay2022study,xu2022contrastive}.

\subsection{Multi-task Learning}
Multi-task learning (MTL) is a framework for learning multiple tasks jointly with shared parameters, rather than learning them independently~\citep{zhang2021survey}. By training multiple tasks simultaneously, MTL can take advantage of shared structures between tasks to improve the generalization performance of each individual task.
Several different MTL architectures have been proposed in the literature, including hard parameter sharing, soft parameter sharing, and task-specific parameter sharing~\citep{ruder2017overview}.
Hard parameter sharing involves using a shared set of parameters for all tasks, while soft parameter sharing allows for different parameter sets for each task, but encourages them to be similar. Task-specific parameter sharing has their own set of task-specific parameters for each task, and lower-level parameters are shared between tasks. This approach allows for greater task-specific adaptation while still leveraging shared information at lower levels. 

One of the major challenges of MTL is balancing task-specific and shared representation. Using more shared representations across tasks yields better generalization and performance across all tasks, but at the expense of task-specific information. Therefore, the choice of MTL architecture and the amount of information shared between tasks depends on the specific problem and task relationship. In this paper, we train our model with a task-specific parameter sharing architecture considering shared information and task relationships.
MTL is applied to various areas such as natural language processing, computer vision. In natural language processing, MTL has been used for tasks such as named entity recognition, part-of-speech tagging, and sentiment analysis~\citep{chen2021multi}. In computer vision, MTL has been used for tasks such as object recognition, object detection, and semantic segmentation~\citep{liu2019end}. 
   
These architectures have demonstrated promising results on several benchmark datasets, and in this paper, we conduct research on time-series anomaly detection using MTL, showing excellent performance for all 17 baselines.

\section{Proposed Method}
In this section, we describe the problem statement of our window-based time series anomaly and precursor-of-anomaly detection in detail and introduce the overall architecture of our proposed method, PAD, followed by our NCDE-based neural network design, its multi-task learning strategy and the knowledge distillation.

\subsection{Problem Statement}

In this section, we define the anomaly detection and the precursor-of-anomaly detection tasks in irregular time series settings --- those problems in the regular time series setting can be similarly defined. 

In our study, we focus on multivariate time series, defined as $\mathbf{x}_{0:T}=\{\mathbf{x}_0,\mathbf{x}_1,...,\mathbf{x}_T\}$, where $T$ is the time-length. An observation at time $t$, denoted $\mathbf{x}_t$, is an $N$-dimensional vector, i.e., $\mathbf{x}_t \in \mathbb{R}^N$. For irregular time-series, the time-interval between two consecutive observations is not a constant --- for regular time-series, the time-interval is fixed. We use a window-based approach where time series $\mathbf{x}_{0:T}$ is divided into non-overlapping windows, i.e., $\mathbf{w}_{i} = [\mathbf{x}_{t^i_0},\mathbf{x}_{t^i_1},...,\mathbf{x}_{t^i_b}]$ where $t^i_j = i  \times b + j - 1$ with a window size of $b$. There are $\lceil \frac{T}{b} \rceil$ windows in total for $\mathbf{x}_{0:T}$. Each window is individually taken as an input to our network for the anomaly or the precursor-of-anomaly detection.

Given an input window $\mathbf{w}_i$, for the anomaly detection, our neural network decides whether $\mathbf{w}_i$ contains abnormal observations or not, i.e., a binary classification of anomaly vs. normal. On the other hand, it is determined whether the very next window $\mathbf{w}_{i+1}$ is likely to contain abnormal observations for the precursor-of-anomaly detection, i.e., yet another binary classification.
\begin{figure} 
\begin{center}
    {\includegraphics[width=\columnwidth]{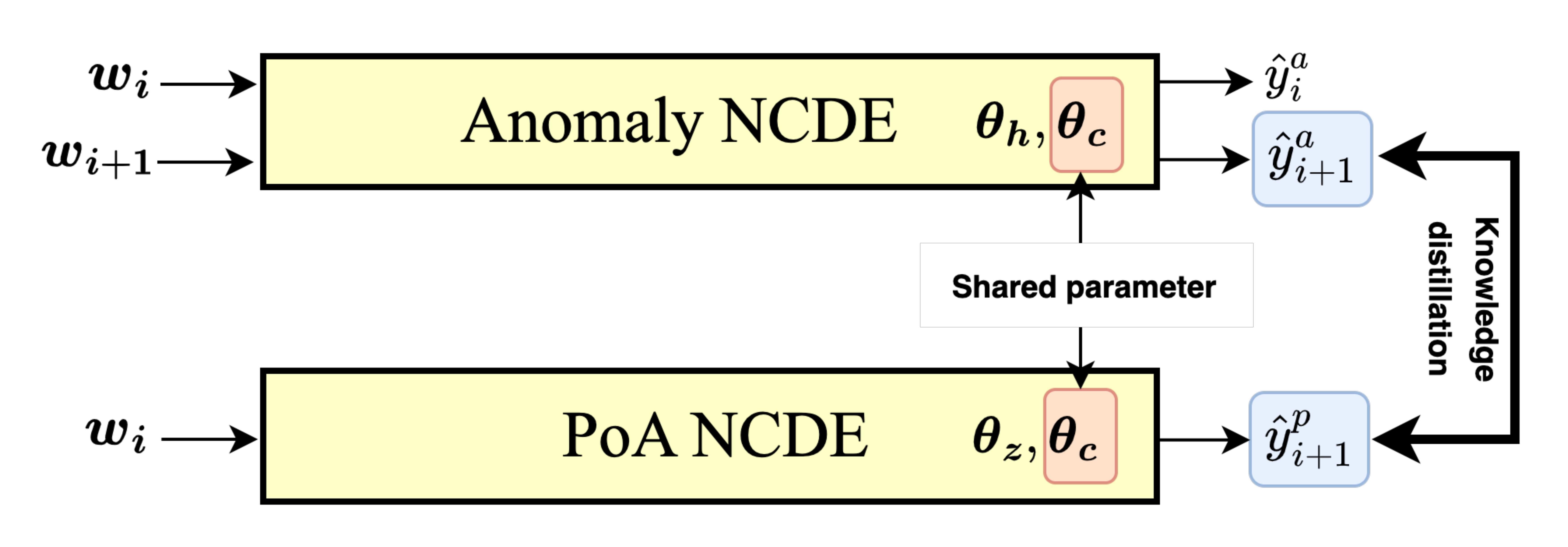}}
    \caption{Overall Architecture}
    \label{fig:overall}
    \end{center}
    \vspace{-1em}
\end{figure}
\subsection{Overall Workflow}

Fig.~\ref{fig:overall} shows the detailed design of our method, PAD. The overall workflow is as follows: 

\begin{enumerate}
    \item For self-supervised learning, we create augmented training data with data augmentation techniques (see Section~\ref{sec:data_augmentation}).
    \item There are two co-evolving NCDE layers which produce the last hidden representations $\mathbf{h}(T)$ and $\mathbf{z}(T)$ in Eq.~\eqref{eq:co-evolving_ncde}.
    \item In the training progress, the anomaly NCDE gets two inputs, $w_i$ for the anomaly detection and $w_{i+1}$ for the PoA detection. 
    \item There are 2 output layers for the anomaly detection and the PoA detection, respectively. These two different tasks are integrated into a single training method via our shared parameter $\theta_c$ for multi-task learning. 
    \item In the training progress, the anomaly NCDE creates the two outputs $\hat{y}^a_i$ and $\hat{y}^a_{i+1}$ for the knowledge distillation.
\end{enumerate}

\subsection{Neural Network Architecture based on Co-evolving NCDEs}
We describe our proposed method based on dual co-evolving NCDEs: one for the anomaly detection and the other for the precursor-of-anomaly (PoA) detection. Given a discrete time series sample $\mathbf{x}_{1:T}$, we create a continuous path $X(t)$ using an interpolation method, which is a pre-processing step of NCDEs. After that, the following co-evolving NCDEs are used to derive the two hidden vectors $\mathbf{h}(T)$ and $\mathbf{z}(T)$:
\begin{align}\begin{split}
    &\mathbf{h}(T) = \mathbf{h}(0) + \int_{0}^{T} f(\mathbf{h}(t);\mathbf{\theta}_f,\mathbf{\theta}_c) \frac{dX(t)}{dt} dt,\\
    &\mathbf{z}(T)=\mathbf{z}(0) + \int_{0}^{T} g(\mathbf{z}(t);\mathbf{\theta}_g,\mathbf{\theta}_c) \frac{dX(t)}{dt} dt,
\end{split}\label{eq:co-evolving_ncde}\end{align} where the two functions (neural networks) $f$ and $g$ have their own parameter and shared parameter $c$ 

For instance, $\mathbf{\theta}_f$ is specific to the function $f$ whereas $\mathbf{\theta}_c$ is a common parameter shared by $f$ and $g$. The exact architectures of $f$ and $g$ are as follows:
\begin{align}
    f(\mathbf{h}(t);\theta_f,\theta_c) &= \underbrace{\rho(\texttt{FC}(\phi(\texttt{FC}(\mathbf{h}(t)))))}_{\theta_f}+ \underbrace{\rho(\texttt{FC}(\phi(\texttt{FC}(\mathbf{h}(t)))))}_{\theta_c},\\
    g(\mathbf{z}(t);\theta_g,\theta_c) &= \underbrace{\rho(\texttt{FC}(\phi(\texttt{FC}(\mathbf{z}(t)))))}_{\theta_g}+ \underbrace{\rho(\texttt{FC}(\phi(\texttt{FC}(\mathbf{z}(t)))))}_{\theta_c}
\end{align}
where $\phi$ means the rectified linear unit (ReLU) and $\rho$ means the hyperbolic tangent. 

Therefore, our proposed architecture falls into the category of \emph{task-specific parameter sharing} of the multi-task learning paradigm. We perform the anomaly detection task and the PoA detection tasks with $\mathbf{h}(T)$ and $\mathbf{z}(T)$, respectively. We ensure that the two NCDEs co-evolve by using the shared parameters $\theta_c$ that allow them to influence each other during the training process. Although those two tasks' goals are ultimately different, those two tasks share common characteristics to some degree, i.e., capturing key patterns from time series. By controlling the sizes of $\mathbf{\theta}_f$, $\mathbf{\theta}_g$, and $\mathbf{\theta}_c $, we can control the degree of their harmonization. After that, we have the following two output layers:
\begin{align}
\hat{y}_{i}^a=&\sigma(\texttt{FC}_{\theta_a}(\mathbf{h}(T))),\textrm{ for the anomaly detection},\\
\hat{y}_{i}^p=&\sigma(\texttt{FC}_{\theta_p}(\mathbf{z}(T))),\textrm{ for the precursor-of-anomaly detection},\label{eq:pd}
\end{align} where each fully-connected (\texttt{FC}) layer is defined for each task and $\sigma$ is the sigmoid activation for binary classification.

When implementing the co-evolving NCDEs, we implement the following augmented state and solve the two initial value problems in Eq.~\eqref{eq:co-evolving_ncde} at the same time with an ODE solver:
\begin{align} \label{eq3}
\frac{d}{dt}{\begin{bmatrix}
  \mathbf{h}(t) \\
  \mathbf{z}(t)
  \end{bmatrix}\!} = {\begin{bmatrix}
  f(\mathbf{h}(t);\mathbf{\theta}_f,\mathbf{\theta}_c) \frac{dX(t)}{dt} \\
  g(\mathbf{z}(t);\mathbf{\theta}_g,\mathbf{\theta}_c) \frac{dX(t)}{dt}
  \end{bmatrix},\!}
\end{align} and 
\begin{align}
{\begin{bmatrix}
  \mathbf{h}(0) \\
  \mathbf{z}(0)
  \end{bmatrix}\!} = {\begin{bmatrix}
  \texttt{FC}_{\theta_\mathbf{h}}(X(0)) \\
  \texttt{FC}_{\theta_\mathbf{z}}(X(0))
  \end{bmatrix}.\!}
\end{align}

\paragraph{\textbf{Well-posedness of the problem:}} The well-posedness of the initial value problem of NCDEs was proved in previous work, such as~\citep{lyons2004differential,NEURIPS2020_4a5876b4} under the condition of Lipschitz continuity, which means that the optimal form of the last hidden state at time $T$ is uniquely defined given an training objective. Our method, PAD, also has this property, as almost all activation functions (e.g. ReLU, Leaky ReLU, Sigmoid, ArcTan, and Softsign) have a Lipschitz constant of 1~\citep{NEURIPS2020_4a5876b4}. Other neural network layers, such as dropout, batch normalization, and pooling methods, also have explicit Lipschitz constant values. Therefore, Lipschitz continuity can be fulfilled in our case.

\subsection{Training Algorithm}\label{sec:training_alg}

\paragraph{\textbf{Loss function:}} We use the cross-entropy (CE) loss to train our model. $L_{KD}$ and $L_a$ mean the CE loss for the knowledge distillation and the anomaly detection, respectively:
\begin{align}
    \begin{split}\label{eq:kd_loss_function}
        L_{KD} &= CE(\hat{y}^a_{i+1},\hat{y}^p_{i+1}), \\
    \end{split}
\end{align} where $\hat{y}^a_{i+1}$ denotes the anomaly NCDE model's output, $\hat{y}^p_{i+1}$ denotes the PoA NCDE model's output and 
\begin{align}
\begin{split}\label{eq:anomaly_loss_function}
        L_a &= CE(\hat{y}^a_i,y_i),\\
    \end{split}
\end{align}
where $y_i$ denotes the ground-truth of anomaly detection. As shown in Eq.~\eqref{eq:kd_loss_function}, we use the cross entropy (CE) loss between $\hat{y}^a_{i+1}$ and $\hat{y}^p_{i+1}$ to distill the knowledge of the anomaly NCDE into the PoA NCDE. 
\paragraph{\textbf{Training with the adjoint method:}}

We train our method using the adjoint sensitivity method ~\citep{NIPS2018_7892,pontryagin1962the,giles2000an,hager2001runge}, which requires a memory of $\mathcal{O}(T+H)$ where $T$ is the integral time domain and $H$ is the size of the NCDE's vector field. This method is used in Lines~\ref{alg:train1} to ~\ref{alg:train2} of Alg.~\ref{alg:train}. However, our framework uses two NCDEs, which increases the required memory to $\mathcal{O}(2T + H_f + H_g)$, where $H_f$ and $H_g$ are the sizes of the vector fields for the two NCDEs, respectively. 

To train them, we need to calculate the gradients of each loss w.r.t. the parameters $\theta_f$, $\theta_g$, and $\theta_c$. In this paragraph, we describe how we can space-efficiently calculate them. The gradient to train each parameter can be defined as follows:

\begin{align}
    \begin{split}
        \nabla_{\mathbf{\theta}_f} L_a &= -\int^{0}_{T}\frac{\partial L_a}{\partial\mathbf{h}(t)}^T\frac{\partial f(\mathbf{h}(t),t;\theta_f,\theta_c)}{\partial \theta_f} \frac{dX(t)}{dt} dt,\\
        \nabla_{\mathbf{\theta}_f} L_{KD} &= -\int^{0}_{T}\frac{\partial L_{KD}}{\partial\mathbf{h}(t)}^T\frac{\partial f(\mathbf{h}(t),t;\theta_f,\theta_c)}{\partial \theta_f}\frac{dX(t)}{dt} dt,\\
        \nabla_{\mathbf{\theta}_g} L_{KD} &= -\int^{0}_{T}\frac{\partial L_{KD}}{\partial\mathbf{z}(t)}^T\frac{\partial g(\mathbf{z}(t),t;\theta_g,\theta_c)}{\partial \theta_g}\frac{dX(t)}{dt} dt,\\
         \nabla_{\mathbf{\theta}_c} L_a + L_{KD}&= -\int^{0}_{T}\frac{\partial L_a + L_{KD}}{\partial\mathbf{h}(t)}^T\frac{\partial f(\mathbf{h}(t),t;\theta_f,\theta_c)}{\partial \theta_c} \frac{dX(t)}{dt} dt \\
        & -\int^{0}_{T}\frac{\partial L_a + L_{KD}}{\partial\mathbf{z}(t)}^T\frac{\partial g(\mathbf{z}(t),t;\theta_g,\theta_c)}{\partial \theta_c} \frac{dX(t)}{dt} dt.
    \end{split}
\end{align}

\begin{algorithm}[t]
\SetAlgoLined
\caption{How to train PAD}\label{alg:train}
\KwIn{Training data $D_{train}$, Validating data $D_{val}$, Maximum iteration number $max\_iter$}
\tcc{$\mathbf{\theta}_{others} = \{\theta_\mathbf{h}, \theta_\mathbf{z}, \mathbf{\theta}_a, \mathbf{\theta}_t, \mathbf{\theta}_p\}$}
Initialize $\mathbf{\theta}_f$, $\mathbf{\theta}_g$, $\mathbf{\theta}_c$, and $\mathbf{\theta}_{others}$;

$k \gets 0$;

\While {$k < max\_iter$}{
    Train $\mathbf{\theta}_f$ and $\mathbf{\theta}_{others}$ with $L_a$;\\ \label{alg:train1}
    Train $\mathbf{\theta}_g$  and $\mathbf{\theta}_{others}$ with $L_{KD}$;\\ \label{alg:train2}
    Train the common parameter $\mathbf{\theta}_c$ with $L_a, L_{KD}$;\\  
    Validate and update the best parameters, $\mathbf{\theta}^*_f$, $\mathbf{\theta}^*_g$,$\mathbf{\theta}^*_c$, $\mathbf{\theta}^*_{others}$ with $D_{val}$\;
    $k \gets k + 1$;
}
\Return $\mathbf{\theta}^*_f$, $\mathbf{\theta}^*_g$, $\mathbf{\theta}^*_c$, and $\mathbf{\theta}^*_{others}$;
\end{algorithm}

\subsection{Data Augmentation Methods for Self-supervised Learning} \label{sec:data_augmentation}
In general, the dataset for anomaly detection do not provide labels for training samples but have labels only for testing samples. Thus, we rely on the self-supervised learning technique. We mainly use re-sampling methods for data augmentation, and we resort to the following steps for augmenting training samples with anomalous patterns:
\begin{enumerate}
    \item Maintain an anomaly ratio $\gamma = \frac{L}{T}$ for each dataset;
    \item The starting points $t_{start}$ of anomaly data points are all designated randomly;
    \item The length vector $\textbf{l}$ of abnormal data  are randomly selected between sequence 100 to sequence 500;
    \item The re-sampled data is randomly selected from the training samples. 
\end{enumerate}

$L$ is the total length of abnormal data points, and $T$ is the total length of data points. A detailed data augmentation algorithm is in Appendix.

\section{Experiments}
In this section, we describe our experimental environments and results. All experiments were conducted in the following software and hardware environments: \textsc{Ubuntu} 18.04 LTS, \textsc{Python} 3.8.13, \textsc{Numpy} 1.21.5, \textsc{Scipy} 1.7.3, \textsc{Matplotlib} 3.3.1, \textsc{PyTorch} 1.7.0, \textsc{CUDA} 11.0, \textsc{NVIDIA} Driver 417.22, i9 CPU, and \textsc{NVIDIA RTX 3090}.

\subsection{Datasets}

\paragraph{\textbf{Mars Science Laboratory:}}
The mars science laboratory (MSL) dataset is also from NASA, which was collected by a spacecraft en route to Mars. This dataset is a publicly available dataset from NASA-designated data centers. It is one of the most widely used dataset for the anomaly detection research due to the clear distinction between pre and post-anomaly recovery. It is comprised of the health check-up data of the instruments during the journey. This dataset is a multivariate time series dataset, and it has 55 dimensions with an anomaly ratio of approximately 10.72\%~\citep{hundman2018detecting}.

\paragraph{\textbf{Secure Water Treatment:}}
The secure water treatment (SWaT) dataset is a reduced representation of a real industrial water treatment plant that produces filtered water. This data set contains important information about effective measures that can be implemented to avoid or mitigate cyberattacks on water treatment facilities. The data set was collected for a total of 11 days, with the first 7 days collected under normal operating conditions and the subsequent 4 days collected under simulated attack scenarios. SWaT has 51 different values in an observation and an anomaly ratio of approximately 11.98\%~\citep{goh2017dataset}.

\paragraph{\textbf{Water Distribution:}}
The water distribution (WADI) data set is compiled from the WADI testbed, an extension of the SWaT testbed. It is measured over 16 days, of which 14 days are measured in the normal state and 2 days are collected in the state of the attack scenario. WADI has 123 different values in an observation and an anomaly ratio of approximately 5.99\%~\citep{mathur2016swat}.


\begin{table*}[t]
\caption{Experimental results for the anomaly detection on 3 regular time-series datasets. P, R, and F1 denote Precision, recall and F1 score as \%, respectively. The best scores are in bold face.}\label{tbl:result}
\renewcommand{\arraystretch}{1.05}
\begin{tabular}{cclllllllll}
\hline
\multicolumn{2}{c}{Datasets}                                 & \multicolumn{3}{c}{MSL} & \multicolumn{3}{c}{SWaT} & \multicolumn{3}{c}{WADI} \\ \hline
\multicolumn{2}{c}{Methods}                                   & P      & R      & F1    & P      & R      & F1    & P      & R      & F1     \\ \hline
\multicolumn{1}{c|}{\multirow{2}{*}{Classical Methods}} &
  \multicolumn{1}{c|}{OCSVM} 
  & 59.78  & 86.87 & 70.82  & 45.39  & 49.22 & 47.23   & 83.93  & 46.42 & 58.64 \\ 
\multicolumn{1}{c|}{} & \multicolumn{1}{c|}{Isolation Forest} 
 & 53.94  & 86.54 & 66.45  & 95.12  & 58.84 & 72.71   & 95.12  & 58.84 & 72.71  \\ \hline
\multicolumn{1}{c|}{\multirow{3}{*}{\begin{tabular}[c]{@{}c@{}}Clustering-based\\ Methods\end{tabular}}} &
  \multicolumn{1}{c|}{Deep-SVDD} 
  & 91.92  & 76.63 & 83.58  & 80.42  & 84.45 & 82.39   & 83.70  & 47.88 & 60.03 \\
\multicolumn{1}{c|}{} & \multicolumn{1}{c|}{ITAD}          
& 69.44  & 84.09 & 76.07  & 63.13  & 52.08 & 57.08   & 92.11  & 58.79 & 70.25  \\
\multicolumn{1}{c|}{} & \multicolumn{1}{c|}{THOC}        
& 88.45  & 90.97 & 89.69  & 98.08  & 79.94 & 88.09   & 86.71  & 92.02 & 89.29  \\ \hline
\multicolumn{1}{c|}{\multirow{3}{*}{\begin{tabular}[c]{@{}c@{}}Density-estimation-based\\ Methods\end{tabular}}} &
\multicolumn{1}{c|}{LOF} 
  & 81.17  & 81.44 & 81.23  & 72.15  & 65.43 & 68.62  & 87.61  & 17.92 & 25.23 \\
\multicolumn{1}{c|}{} & \multicolumn{1}{c|}{DAGMM}          
 &89.60  & 63.93 & 74.62  & 89.92  & 57.84 & 70.40   & 46.95  & 66.59 & 55.07  \\
\multicolumn{1}{c|}{} & \multicolumn{1}{c|}{MMPCACD}        
& 81.42  & 61.31 & 69.95  & 82.52  & 68.29 & 74.73   & 88.61  & 75.84 & 81.73  \\ \hline
\multicolumn{1}{c|}{\multirow{8}{*}{\begin{tabular}[c]{@{}c@{}}Reconstruction-based\\ Methods\end{tabular}}} &
  \multicolumn{1}{c|}{VAR} 
& 74.68  & 81.42 & 77.90  & 81.59  & 60.29 & 69.34   & 83.97 & 49.35 & 61.31\\
\multicolumn{1}{c|}{} & \multicolumn{1}{c|}{LSTM}            
& 85.45  & 82.50 & 83.95  & 86.15  & 83.27 & 84.69   & 81.43  & 84.82 & 83.06  \\
\multicolumn{1}{c|}{} & \multicolumn{1}{c|}{CL-MPPCA}        
& 73.71  & 88.54 & 80.44  & 76.78  & 81.50 & 79.07   & 70.96  & 75.21 & 72.86  \\ 
 \multicolumn{1}{c|}{} &\multicolumn{1}{c|}{LSTM-VAE} 
& 85.49  & 79.94 & 82.62  & 76.00  & 89.50 & 82.20   & 98.97  & 63.77 & 77.56 \\
\multicolumn{1}{c|}{} & \multicolumn{1}{c|}{BeatGAN}            
& 89.75  & 85.42 & 87.53  & 64.01  & 87.46 & 73.92   & 70.48  & 72.26 & 71.34  \\
\multicolumn{1}{c|}{} & \multicolumn{1}{c|}{OmniAnomay}        
& 89.02  & 86.37 & 87.67  & 81.42  & 84.30 & 82.83   & 98.25  & 64.97 & 78.22  \\
\multicolumn{1}{c|}{} & \multicolumn{1}{c|}{USAD}   
& 89.36  & 92.92 & 91.05  & 98.51  & 66.18 & 79.17   & 99.47  & 13.18 & 23.28  \\
\multicolumn{1}{c|}{} & \multicolumn{1}{c|}{InterFusion}            
& 81.28  & 92.70 & 86.62  & 80.59  & 85.58 & 83.01   & 90.30  & 92.67 & 91.02  \\
\multicolumn{1}{c|}{} & \multicolumn{1}{c|}{Anomaly Transformer}  
& 91.82  & 91.23 & 91.53  & 87.32  & 85.50 & 86.40   & 60.86  & 77.86 & 68.32  \\\hline
\multicolumn{1}{c|}{\multirow{1}{*}{Ours}} &
  \multicolumn{1}{c|}{\textbf{PAD} (Anomaly)}
  & 94.13  & 94.71 &\textbf{92.56}  & 94.02  & 93.53 & \textbf{93.04}   & 90.84  & 95.31 & \textbf{93.02}\\  \hline
\end{tabular}
\end{table*}

\subsection{Experimental Settings}

\subsubsection{Hyperparameters}
We list all the detailed hyperparameter setting for baselines and our method in Appendix. 

For reproducibility, we report the following best hyperparameters for our method: 
\begin{enumerate}
    \item In MSL, we train for $300$ epochs, a learning rate of $1.0 \times e^{-2}$, a weight decay of $1.0 \times e^{-4}$, and the hidden size of $\theta_f, \theta_g$, and $\theta_c$ is $256,512,256$, respectively. Among $256$ windows, we detect the window in which abnormal data points exist. The length of each window was set to $30$, and the length of the predicted window in precursor anomaly detection was set to $10$.
    \item In SWaT, we train for $300$ epochs, a learning rate of $1.0 \times e^{-2}$, a weight decay of $1.0 \times e^{-4}$, and the hidden size of $\theta_f, \theta_g$, and $\theta_c$ is $128,64,64$, respectively. Among $256$ windows, we detect the window in which abnormal data points exist. The length of each window was set to $60$, and the length of the predicted window in precursor anomaly detection was set to $20$.
    \item In WADI, we train for $300$ epochs, a learning rate of $1.0\times e^{-2}$, a weight decay of $1.0\times e^{-5}$, and the hidden size of $\theta_f, \theta_g$, and $\theta_c$ is $128,128,256$, respectively. Among $256$ windows, we detect the window in which abnormal data points exist. The length of each window was set to $100$, and the length of the predicted window in precursor anomaly detection was set to $30$.
\end{enumerate}
\subsubsection{Baselines}
We list all the detailed description about baselines in Appendix.
We compare our model with the following 16 baselines of 4 categories, including not only traditional methods but also state-of-the-art deep learning-based models as follows:
\begin{enumerate}
    \item The classical method category includes OCSVM~\citep{tax2004support} and Isolation Forest~\citep{liu2008isolation}.
    \item The clustering-based methods category includes Deep-SVDD 
    \citep{ruff2019deep}, ITAD~\citep{shin2020itad}, and THOC~\citep{shen2020timeseries}.
    \item The density-estimation-based methods category has LOF~\citep{breunig2000lof}, DAGMM~\citep{zong2018deep}, and MMPCACD~\citep{yairi2017data}.
    \item The reconstruction-based methods includes VAR~\citep{Anderson1976TimeSeries2E}, LSTM \citep{hundman2018detecting}, CL-MPPCA~\citep{tariq2019detecting}, LSTM-VAE~\citep{park2018multimodal}, BeatGAN \citep{zhou2019beatgan}, OmniAnomaly~\citep{su2019robust}, USAD~\citep{audibert2020usad}, InterFusion~\citep{li2021multivariate}, and Anomaly Transformer~\citep{xu2021anomaly}.
\end{enumerate}

\subsection{Experimental Results on Anomaly Detection}
We introduce our experimental results for the anomaly detection with the following 3 datasets: MSL, SWaT, and WADI. Evaluating the performance with these datasets proves the competence of our model in various fields. We use the Precision, Recall, and F1-score.

\begin{table*}[t]
\caption{Experimental results for the anomaly detection on 3 irregular time-series datasets.}\label{tbl:result_irr}
\setlength{\tabcolsep}{7pt}
\begin{tabular}{cclllllllll}
\hline
\multicolumn{2}{c}{Dropping ratio}                                 & \multicolumn{3}{c}{30\% dropped} & \multicolumn{3}{c}{50\% dropped} & \multicolumn{3}{c}{70\% dropped} \\ \hline
Datasets &    Methods            & P      & R      & F1    & P      & R      & F1    & P      & R      & F1     \\ \hline
\multicolumn{1}{c|}{\multirow{5}{*}{MSL}} &
\multicolumn{1}{c|}{Isolation Forest} 
   & 89.81  & 41.59 & 52.53  &  86.97  & 39.38 & 50.74   & 88.03  & 29.27 & 38.33 \\ 
\multicolumn{1}{c|}{} & \multicolumn{1}{c|}{LOF} 
 & 78.26  & 78.71 & 78.39  & 74.86  & 75.39 & 74.95   & 73.53  & 72.59 & 72.99 \\ 
 \multicolumn{1}{c|}{} & \multicolumn{1}{c|}{USAD} 
 & 88.07  & 84.04 & 85.92   & 87.49  & 68.73 & 76.26   & 87.49  & 74.92 & 80.33 \\ 
  \multicolumn{1}{c|}{} & \multicolumn{1}{c|}{Anomaly Transformer} 
  & 90.49  & 85.42 & 87.89  & 91.26  & 86.40 & 88.76 & 91.75  & 90.47 & 91.10 \\  
  \multicolumn{1}{c|}{} & \multicolumn{1}{c|}{\textbf{PAD} (Anomaly)} 
 & 89.09  & 94.39 & \textbf{91.66}  & 94.92  & 94.63 & \textbf{92.24} & 91.09  & 94.22 & \textbf{91.87} \\\hline
\multicolumn{1}{c|}{\multirow{5}{*}{\begin{tabular}[c]{@{}c@{}}SWaT\end{tabular}}} &
\multicolumn{1}{c|}{Isolation Forest} 
  & 69.55  & 49.79 & 55.06   & 69.94  & 35.00 & 37.16    & 67.89  & 24.37 & 18.83 \\ 
\multicolumn{1}{c|}{} & \multicolumn{1}{c|}{LOF} 
 & 70.85  & 21.87 & 12.77   & 71.20  & 25.30 & 20.32    & 69.34  & 23.24 & 16.39 \\ 
 \multicolumn{1}{c|}{} & \multicolumn{1}{c|}{USAD} 
 & 87.36  & 52.64 & 64.87   & 84.58  & 38.44 & 50.42    & 87.96  & 47.19 & 59.91  \\ 
 \multicolumn{1}{c|}{} & \multicolumn{1}{c|}{Anomaly Transformer} 
  & 94.48  & 77.61 & 85.22    & 93.33  & 84.14 & 88.50     & 93.23  & 82.81 & 87.71  \\  
  \multicolumn{1}{c|}{} & \multicolumn{1}{c|}{\textbf{PAD} (Anomaly)}
 & 93.65  & 93.27 & \textbf{92.76}   & 93.89  & 93.53 & \textbf{93.07}    & 92.36  & 92.47 & \textbf{92.08}  \\\hline
\multicolumn{1}{c|}{\multirow{5}{*}{\begin{tabular}[c]{@{}c@{}}WADI\end{tabular}}} &
\multicolumn{1}{c|}{Isolation Forest} 
  & 92.19  & 48.48 & 62.33   & 91.77  & 26.35 & 38.03    & 92.89  & 12.15 & 16.23  \\ 
\multicolumn{1}{c|}{} & \multicolumn{1}{c|}{LOF} 
& 86.42  & 16.35 & 21.42   & 85.78  & 12.70 & 15.88    & 91.12  & 10.26 & 13.33  \\ 
 \multicolumn{1}{c|}{} & \multicolumn{1}{c|}{USAD} 
 & 84.98 & 15.70 & 20.60   & 83.78  & 15.95 & 21.30    & 75.89  & 13.19 & 20.09  \\ 
 \multicolumn{1}{c|}{} & \multicolumn{1}{c|}{Anomaly Transformer} 
 & 65.73  & 100.0 & 79.32   & 62.93  & 84.67 & 72.20    & 61.17  & 77.86 & 68.52  \\ 
  \multicolumn{1}{c|}{} & \multicolumn{1}{c|}{\textbf{PAD} (Anomaly)}
 & 95.69  & 95.49 & \textbf{93.44}   & 91.21  & 93.06 & \textbf{92.10}   & 90.85  & 89.41 & \textbf{90.12}  \\\hline
\end{tabular}
\end{table*}

\begin{table}[t] 
\setlength{\tabcolsep}{1.3pt}
\caption{Experimental results for the precursor-of-anomaly detection on 3 regular time-series datasets.}\label{tbl:result_p}
\begin{tabular}{llllllllll}
\hline
Dataset                         & \multicolumn{3}{l}{MSL} & \multicolumn{3}{l}{SWaT} & \multicolumn{3}{l}{WADI}        \\ \hline

Methods                          & P      & R     & F1     & P      & R     & F1      & P      & R     & F1 \\ \hline
LSTM                            & 90.42  & 59.53 & 70.49  & 64.96  & 80.64 & 71.94   & 92.87  & 55.38 & 68.02                         \\ 
LSTM-VAE                        & 69.54  & 83.05 & 75.65   & 75.64  & 55.10  & 61.11   & 84.32  & 91.83 & 87.91   \\ 

USAD                            & 90.78  & 95.27 & 92.97  & 84.08  & 75.07 & 77.42   & 94.63  & 44.10 & 57.40                      \\ \hline
\textbf{PAD} (PoA)          & 91.41  & 95.61 & \textbf{93.46}  & 93.31  & 93.47 & \textbf{93.32}   & 92.71  & 92.71 & \textbf{92.71}\\ 
\hline
\end{tabular} 
\end{table}



\begin{table*}[t]
\caption{Experimental results for the precursor-of-anomaly detection on the 3 irregular time-series datasets.}\label{tbl:result_irrp}
\setlength{\tabcolsep}{7pt}
\begin{tabular}{cclllllllll}
\hline
\multicolumn{2}{c}{Dropping ratio}                                 & \multicolumn{3}{c}{30\% dropped} & \multicolumn{3}{c}{50\% dropped} & \multicolumn{3}{c}{70\% dropped} \\ \hline
Datasets &    Methods            & P      & R      & F1    & P      & R      & F1    & P      & R      & F1     \\ \hline
\multicolumn{1}{c|}{\multirow{5}{*}{MSL}} &
\multicolumn{1}{c|}{LSTM} 
  & 66.88  & 81.16 & 73.29  & 90.06  & 50.90 & 63.28   & 90.09  & 48.85 & 61.41  \\ 
\multicolumn{1}{c|}{} & \multicolumn{1}{c|}{LSTM-VAE} 
 & 66.96  & 81.75 & 73.58  & 90.22  & 51.29 & 63.60   & 90.12  & 48.78 & 61.34 \\ 
 \multicolumn{1}{c|}{} & \multicolumn{1}{c|}{USAD} 
 & 92.09  & 88.27 & 90.03  & 91.54  & 78.33 & 84.04  & 91.17  & 69.54 & 78.27 \\
   \multicolumn{1}{c|}{} & \multicolumn{1}{c|}{\textbf{PAD} (PoA)}
 & 91.38  & 95.04 & \textbf{93.17}   & 91.40  & 95.52 & \textbf{93.42}    & 93.50  & 95.52 & \textbf{93.84}\\\hline
\multicolumn{1}{c|}{\multirow{5}{*}{\begin{tabular}[c]{@{}c@{}}SWaT\end{tabular}}} &
\multicolumn{1}{c|}{LSTM} 
  & 83.57  & 68.87 & 72.11   & 80.15  & 54.13 & 58.21    & 74.56  & 30.27 & 27.99 \\ 
\multicolumn{1}{c|}{} & \multicolumn{1}{c|}{LSTM-VAE} 
 & 81.98  & 56.27 & 60.22   & 71.34  & 35.80 & 37.61    & 74.94  & 30.80 & 28.80 \\ 
 \multicolumn{1}{c|}{} & \multicolumn{1}{c|}{USAD} 
 & 84.12  & 75.20 & 77.54   & 83.58  & 71.80 & 74.65    & 83.15  & 66.60 & 70.09 \\ 
   \multicolumn{1}{c|}{} & \multicolumn{1}{c|}{\textbf{PAD} (PoA)}
 & 85.50  & 77.40 & \textbf{79.47}   & 84.15  & 77.67 & \textbf{79.53}   & 90.85  & 91.07 & \textbf{90.93}  \\\hline
\multicolumn{1}{c|}{\multirow{5}{*}{\begin{tabular}[c]{@{}c@{}}WADI\end{tabular}}} &
\multicolumn{1}{c|}{LSTM} 
  & 91.15  & 46.01 & 59.95   & 91.28  & 25.52 & 36.68    & 92.97  & 18.58 & 26.29 \\ 
\multicolumn{1}{c|}{} & \multicolumn{1}{c|}{LSTM-VAE} 
 & 89.46  & 45.66 & 59.17   & 91.55  & 26.91 & 38.47    & 93.11  & 19.27 & 27.32 \\ 
 \multicolumn{1}{c|}{} & 
 \multicolumn{1}{c|}{USAD} 
 & 94.87  & 39.93 & 53.05   & 94.62  &34.38 & 46.95    & 94.87  & 28.30 & 39.49 \\ 
   \multicolumn{1}{c|}{} & \multicolumn{1}{c|}{\textbf{PAD} (PoA)} 
& 92.15  & 95.49 & \textbf{93.79}   & 92.17  & 96.01 & \textbf{94.05}    & 92.15  & 95.31 & \textbf{93.70}\\\hline
\end{tabular}
\end{table*}

\begin{figure*}[!ht]
    \centering
     \subfigure[MSL]{\includegraphics[width=0.54\columnwidth]{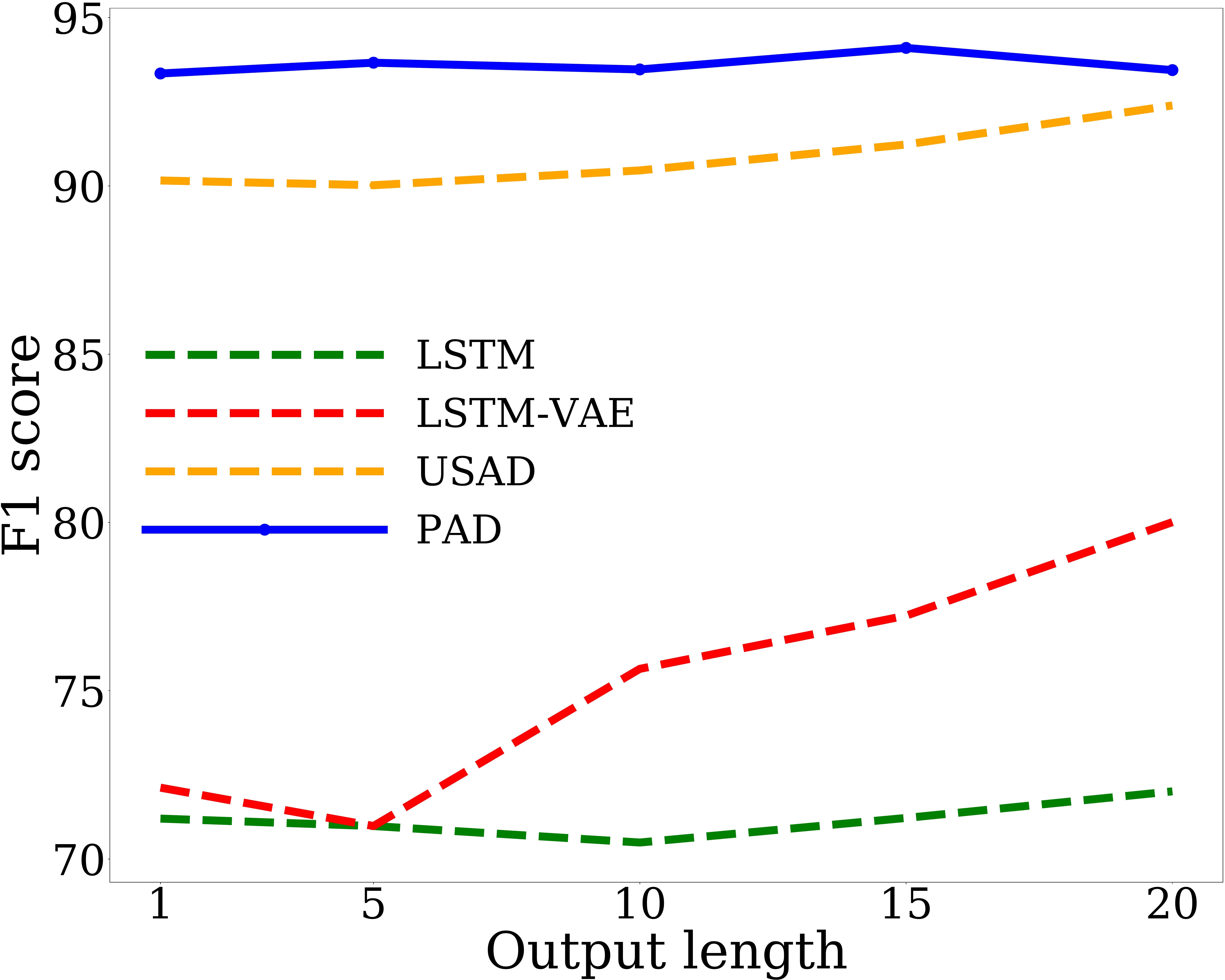}} \hfill
    \subfigure[SWaT]{\includegraphics[width=0.54\columnwidth]{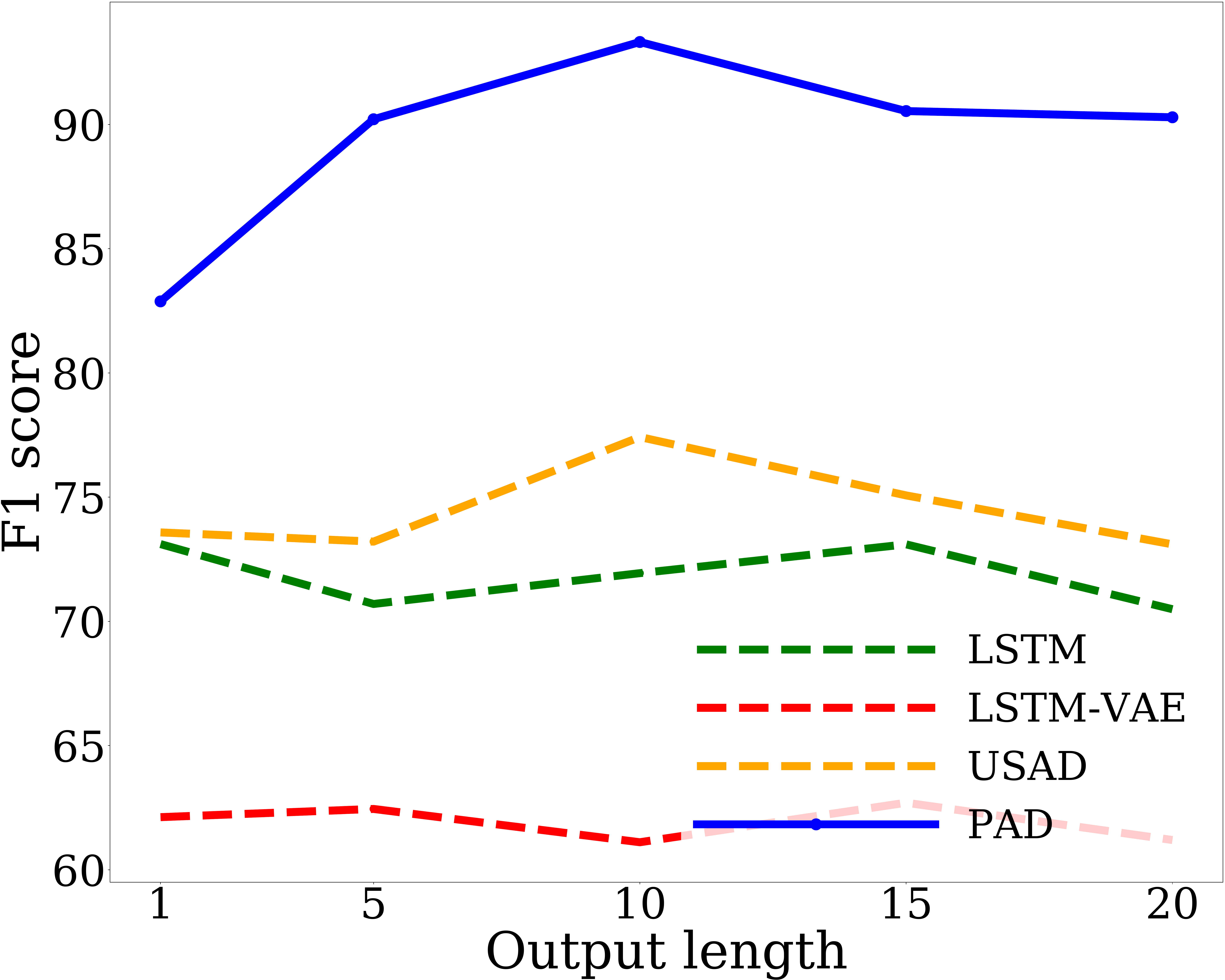}} \hfill
    \subfigure[WADI]{\includegraphics[width=0.54\columnwidth]{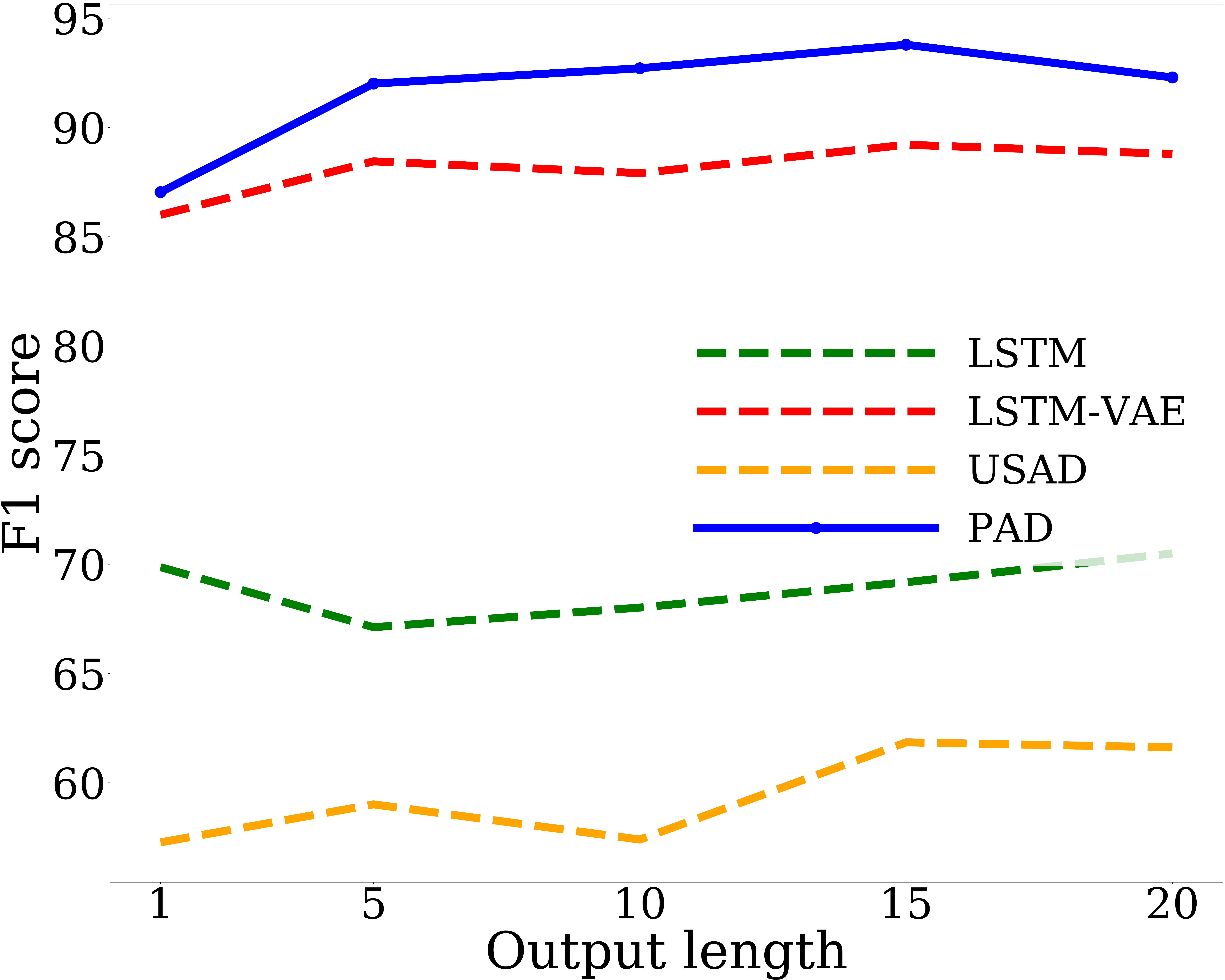}} 
    \caption{F1 score by varying the output length on 3 datasets.}
    \label{fig:experimental_output}
\end{figure*}

\begin{figure}[t]
    \centering
    \subfigure[MSL]{\includegraphics[width=0.45\columnwidth]{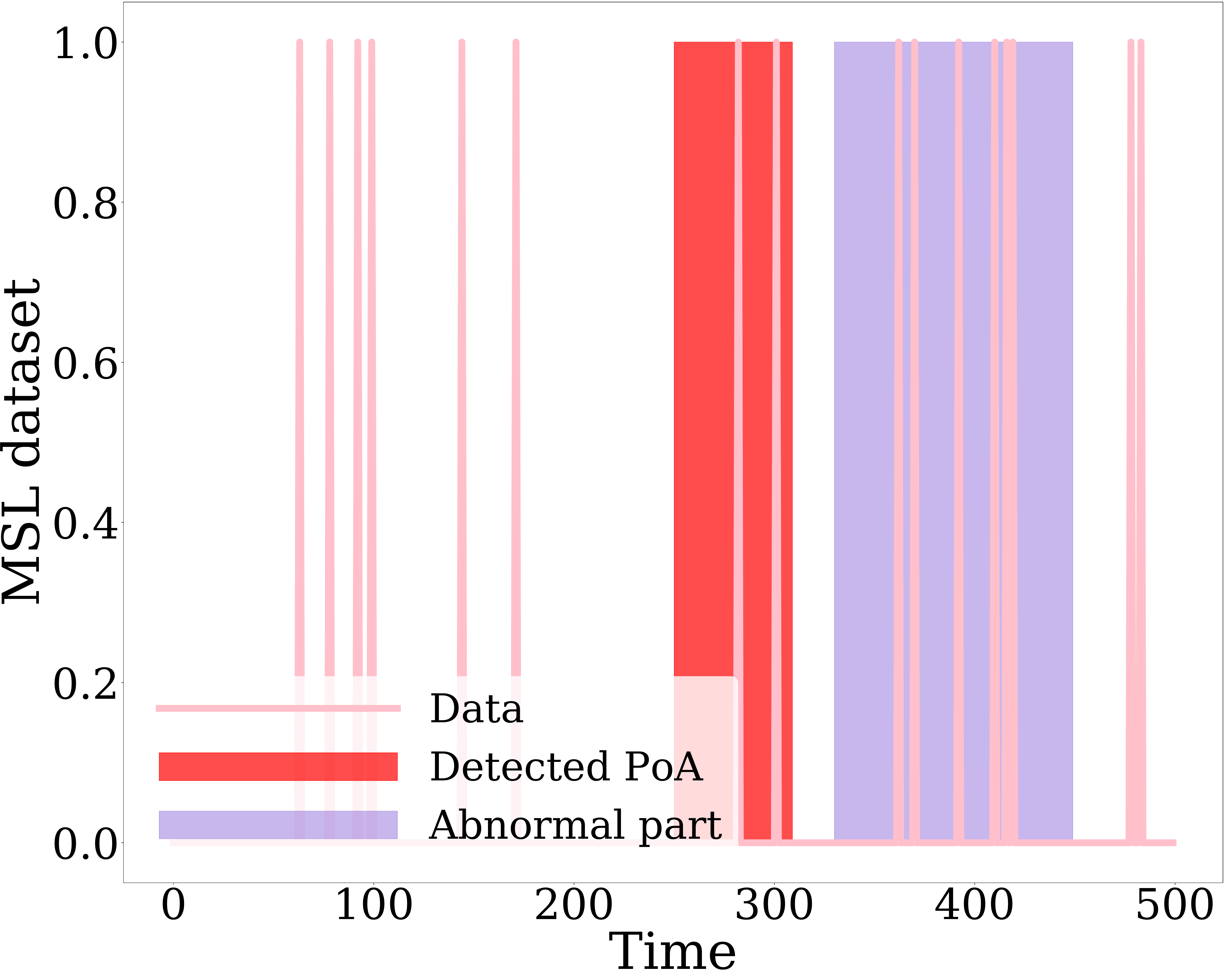}} \hfill
    \subfigure[MSL]{\includegraphics[width=0.45\columnwidth]{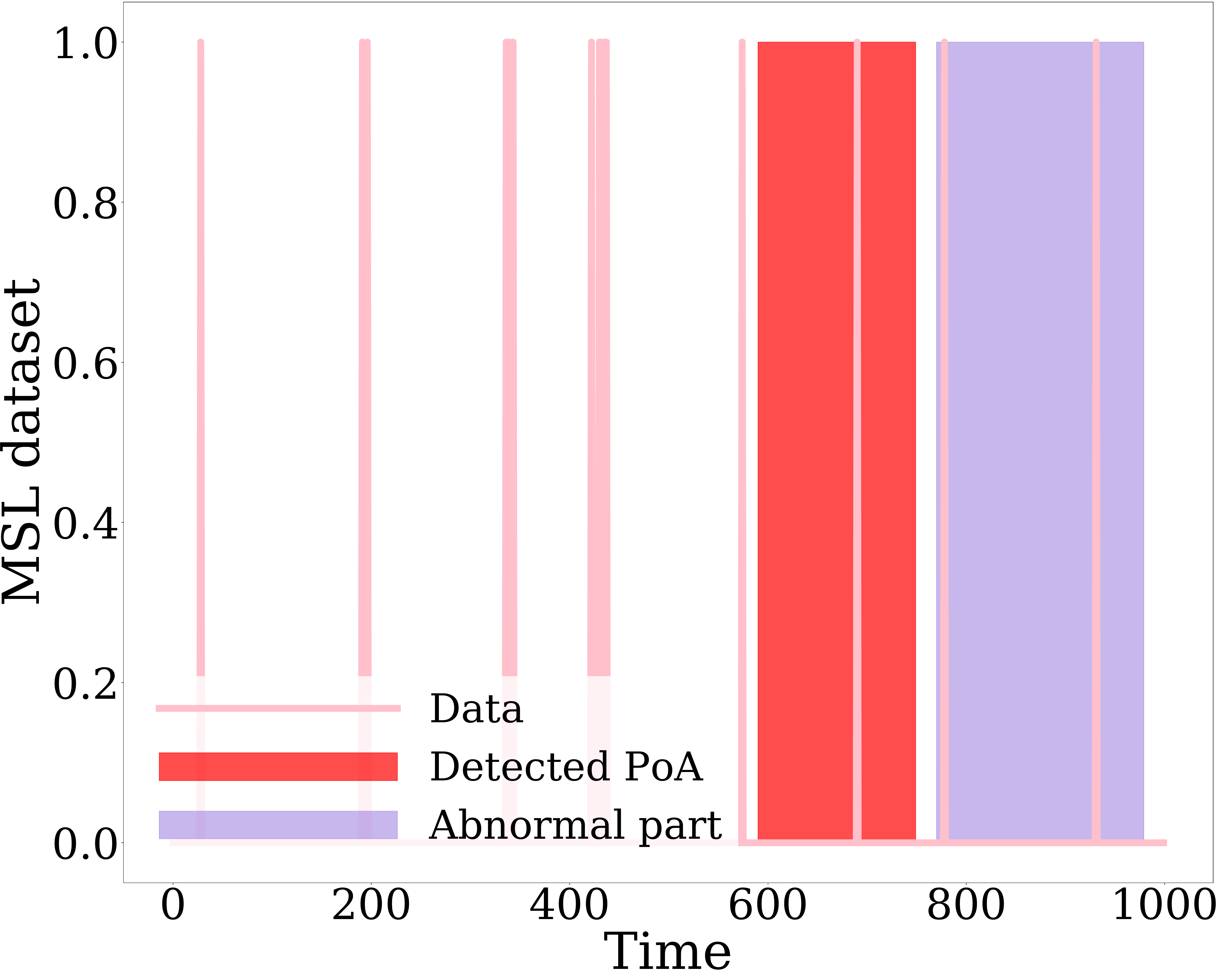}} \\
     \subfigure[SWaT]{\includegraphics[width=0.45\columnwidth]{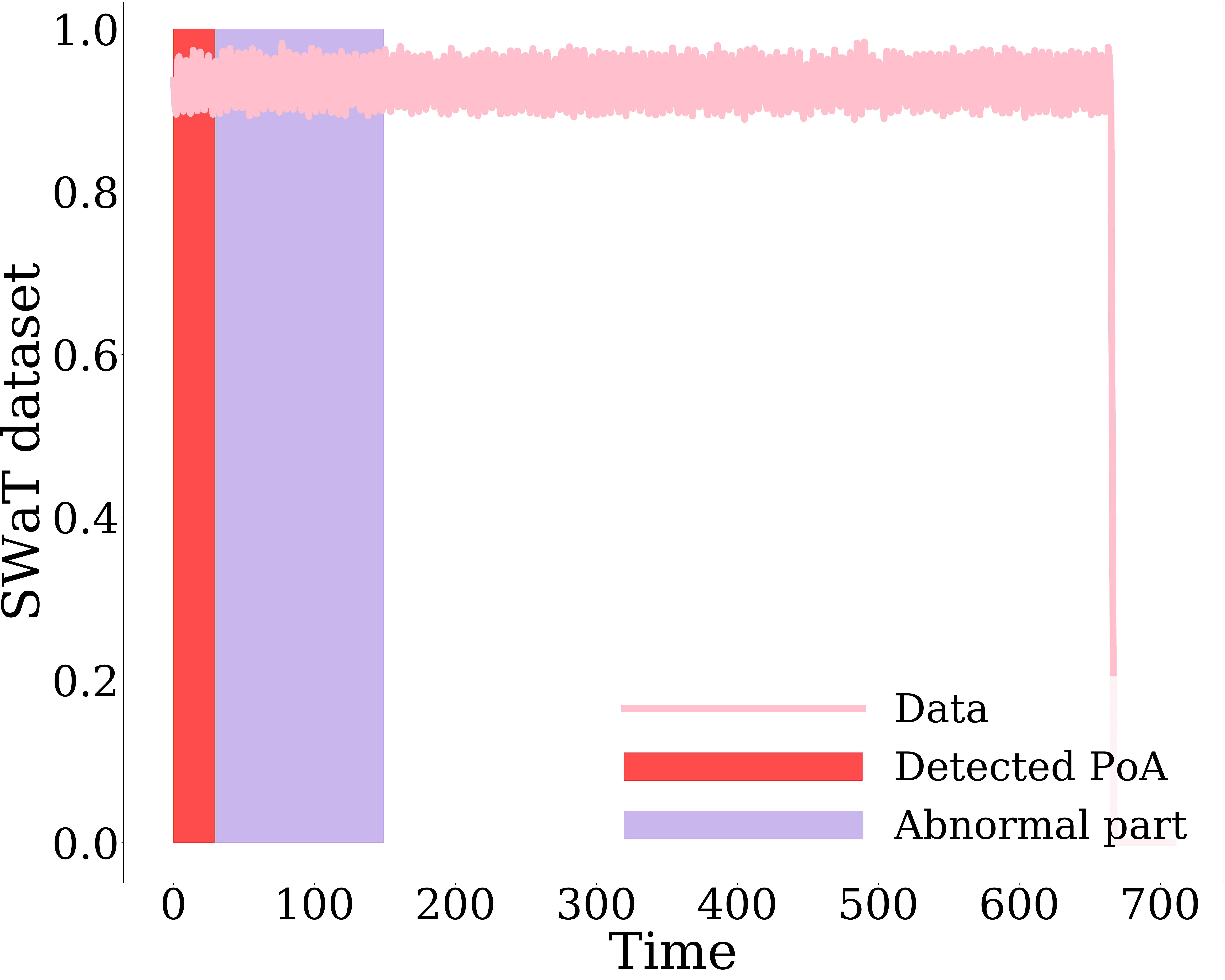}} \hfill
     \subfigure[SWaT]{\includegraphics[width=0.45\columnwidth]{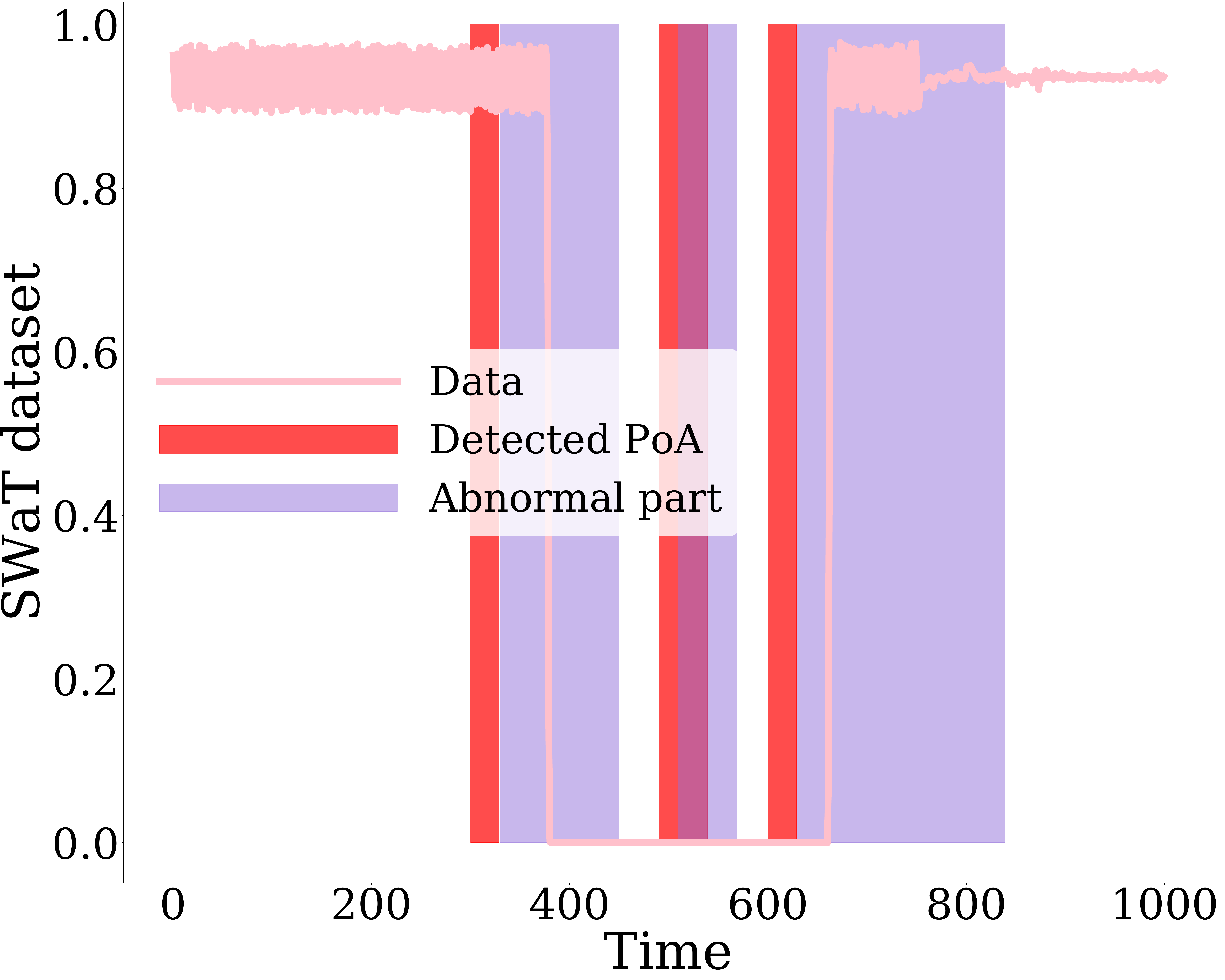}} \\
    \subfigure[WADI]{\includegraphics[width=0.45\columnwidth]{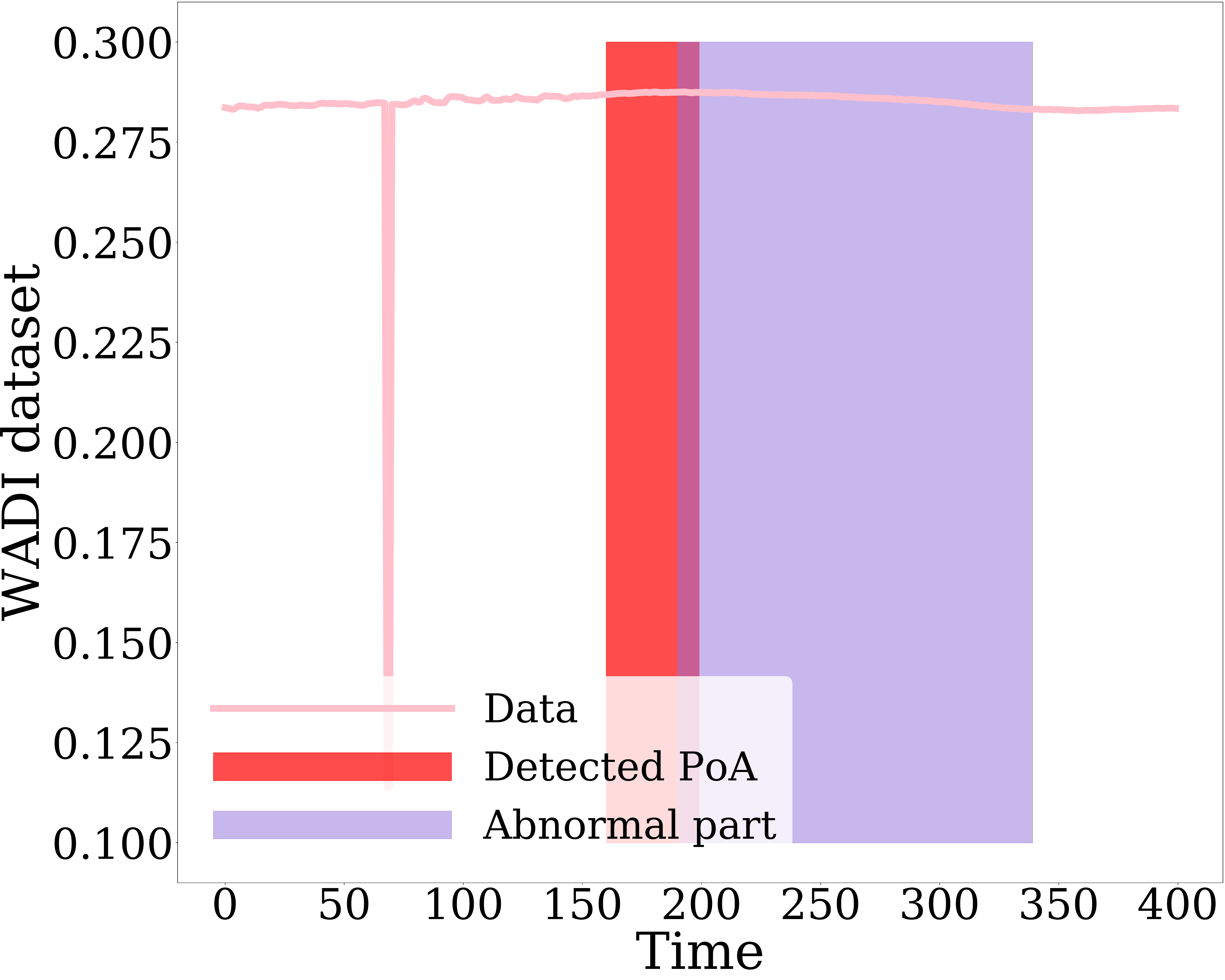}} \hfill
    \subfigure[WADI]{\includegraphics[width=0.45\columnwidth]{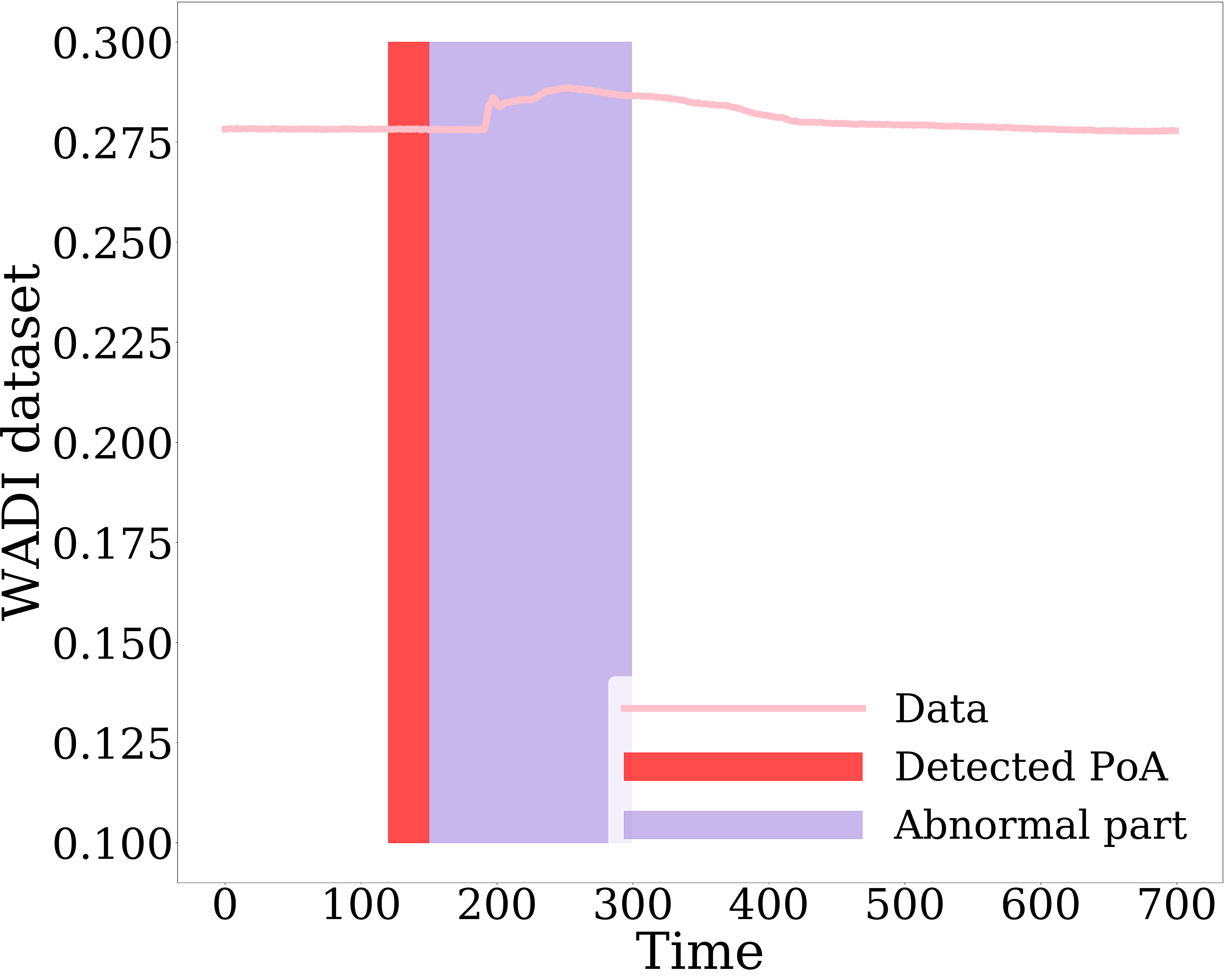}} \\
    \caption{The anomaly detection and the precursor-of-anomaly detection results on 3 datasets.}
    \label{fig:experimental_result_vis}
\end{figure}

\begin{table}[] 
\caption{Ablation study for PAD in a regular setting.}\label{tbl:abl}
\setlength{\tabcolsep}{1.3pt}
\renewcommand{\arraystretch}{1.1}
\begin{tabular}{llllllllll}
\hline
Dataset                         & \multicolumn{3}{l}{MSL} & \multicolumn{3}{l}{SWaT} & \multicolumn{3}{l}{WADI}        \\ \hline
Metric                          & P      & R     & F1     & P      & R     & F1      & P      & R     & F1 \\ \hline
Type (i)            & 94.92  & 94.63 & 92.24  & 94.13  & 93.67 & 93.19    & 90.45  & 87.15 & 88.77                   \\ 
Type (ii)            & 89.74  & 76.65 & 82.30   & 74.94  & 80.07 & 72.11    & 94.57  & 86.98 & 89.97                   \\ \hline
\textbf{PAD}                 & 94.13  & 94.71 &\textbf{92.56}  & 94.02  & 93.53 & \textbf{93.04}   & 90.84  & 95.31 & \textbf{93.02}\\ 
\hline
\end{tabular} 
\end{table}

\begin{table}[] 
\caption{Ablation study for the Precursor-of-Anomaly PAD in a regular setting.}\label{tbl:abl_p}
\setlength{\tabcolsep}{1.3pt}
\renewcommand{\arraystretch}{1.1}
\begin{tabular}{llllllllll}
\hline
Dataset                         & \multicolumn{3}{l}{ MSL} & \multicolumn{3}{l}{SWaT} & \multicolumn{3}{l}{WADI}        \\ \hline
Metric                          & P      & R     & F1     & P      & R     & F1      & P      & R     & F1 \\ \hline
Type (i)     & 91.84  & 88.93 & 90.33   & 90.82  & 90.67 & 90.74    & 91.57  & 82.47 & 86.78 \\
Type (ii)      & 91.37  & 89.18 & 90.26   & 64.96  & 80.60 & 71.94    & 93.12  & 89.06 & 90.95 \\   \hline
\textbf{PAD} (PoA)   & 91.41  & 95.61 & \textbf{93.46}  & 93.31  & 93.47 & \textbf{93.32}   & 92.71  & 92.71 & \textbf{92.71}\\  
\hline
\end{tabular} 
\end{table}

\subsubsection{Experimental Results on Regular Time Series}
Table~\ref{tbl:result} summarizes the results on the three datasets. The anomaly detection with MSL is one of the most widely used benchmark experiments. Our method, PAD, shows the best F1-score. For this dataset, all classical methods are inferior to other baselines. For SWaT, our experimental results are in Table~\ref{tbl:result}. As summarized, the classical methods are inferior to other baselines. However, unlike in WADI, all baselines except them show similar results. Our method, PAD, shows the best F1-score.
For WADI, among the reconstruction-based methods, InterFusion shows the second-best performance. Since this dataset has the smallest anomaly ratio among the three datasets, classical methods and clustering-based methods are not suitable.

\subsubsection{Experimental Results on Irregular Time Series}
Table~\ref{tbl:result_irr} summarizes the results on irregular time series. In order to create challenging irregular environments, we randomly remove 30\%, 50\% and 70\% of the observations in each sequence. Therefore, this is basically an irregular time series anomaly detection. We compare our method, PAD, with the 4 baselines, Isolation Forest, LOF, USAD, and Anomaly Transformer --- other baselines are not defined for irregular time series. In addition, it is expected that the presence of more than 30\% of missing values in time series causes poor performance for many baselines because it is difficult to understand the highly missing input sequence. In MSL, USAD and Anomaly Transformer shows the reasonable results and also maintains an F1-score around 80\% across all the dropping ratio settings. For WADI, all baselines show poor performance when the missing rate is 70\%. Surprisingly, our method, PAD, performs not significantly differently from the regular anomaly detection experiments. Our method maintains good performance in the irregular time series setting as well because PAD uses a hidden representation controlled by the continuous path $X(t)$ at every time $t$. Additionally, our method maintains an F1 score larger than 90\% for all the dropping ratios.

\subsection{Experimental Results on the precursor-of-Anomaly Detection}

In Table.~\ref{tbl:result_p}, we introduce our experimental results for the precursor-of-anomaly detection. In order to compare the performance of our method in the precursor-of-anomaly detection task newly proposed in this paper, among the baselines performed in regular time series anomaly detection (cf. Table~\ref{tbl:result}), we selected reconstruction-based methods that allow PoA experimental setting. Therefore, we select the 3 baselines (LSTM, LSTM-VAE, and USAD) that showed good performance in reconstruction-based methods. 

\subsubsection{Experimental Results on Regular Time Series}
As shown in Table~\ref{tbl:result_p}, USAD shows reasonable performance among the 3 baselines. Especially, in MSL dataset, USAD shows a similar performance to Table~\ref{tbl:result}. Our newly proposed the precursor-of-anomaly task requires predicting patterns or features of future data from input data. Therefore, the reconstruction-based method seems to have shown good performance. However, our method, PAD, shows the best performance in all the 3 datasets. Fig.~\ref{fig:experimental_result_vis} shows the visualization of experimental results on the anomaly detection and the precursor-of-anomaly detection on all the 3 datasets. In Fig.~\ref{fig:experimental_result_vis}, the part highlighted in purple is the ground truth of the anomalies, the part highlighted in red is the result of PoA detected by PAD. As shown in Fig.~\ref{fig:experimental_result_vis}, our method correctly predicts the precursor-of-anomalies (highlighted in red) before the abnormal parts (highlighted in purple) occur.

\subsubsection{Experimental Results on Irregular Time Series}
Table~\ref{tbl:result_irrp} shows the experimental result on the irregular datasets. Among the baselines, USAD has many differences in experimental results depending on the experimental environment. For example, in the WADI dataset, which has a small anomaly ratio(5.99\%) among the other datasets, it shows poor performance, and in the MSL data set, USAD shows the second-best performance, but the performance deteriorates as the dropping ratio increases. However, our method, PAD, clearly shows the best F1-score for all dropping ratios and all 3 datasets. One outstanding point in our model is that the F1-score is not greatly influenced by the dropping ratio.
Consequently, all these results prove that our model shows state-of-the-art performance in both the anomaly and the precursor-of-anomaly detection.

\section{Ablation and Sensitivity Studies}
\subsection{Ablation Study on Multi-task Learning}\label{abl:mtl}
To prove the efficacy of our multi-task learning on the anomaly and the precursor-of-anomaly detection, we conduct ablation studies. There are 2 tasks in our multi-task learning: the anomaly detection, and the precursor-of-anomaly detection tasks. We remove one task to build an ablation model. For the ablation study on anomaly detection, there are 2 ablation models in terms of the multi-task learning setting: i) without the precursor-of-anomaly detection, and ii) with anomaly detection only. For the ablation study on the precursor-of-anomaly detection, 2 ablation models are defined in the exactly same way. Table~\ref{tbl:abl} and Table~\ref{tbl:abl_p} show the results of the ablation studies in the regular time series setting for `PAD (anomaly)' and `PAD (PoA),' respectively. When we remove task(anomaly detection or PoA detection) from the multi-task learning, there is notable degradation in performance. Therefore, our multi-task learning design is required for good performance in both the anomaly and the precursor-of-anomaly detection.

\subsection{Sensitivity to Output Sequence Length}
We also compare our method with USAD, OmniAnomaly, and THOC by varying the length of output of the precursor-of-anomaly detection, during the multi-task learning process. After fixing the input length of MSL, SWaT, and WADI to $30$, we vary the output length in $\{1,5,10,15,20\}$. As shown in Fig.~\ref{fig:experimental_output}, our proposed method consistently outperforms others. As the output length increases, it becomes more difficult to predict in general, but PAD shows excellent performance regardless of the output length. In the MSL and WADI datasets, most baselines show similar performances regardless of the output length. However, in SWaT, there is a performance difference according to the output length, and this phenomenon appears similarly for the baselines.

\section{Conclusion}
Recently, many studies have been conducted on time series anomaly detection. However, most of the methods have been conducted only for existing anomaly detection methods. In this paper, we first propose a task called the precursor-of-anomaly (PoA) detection. We define PoA detection as the task of predicting future anomaly detection. This study is necessary in that many risks can be minimized by detecting risks in advance in the real world. In addition, we combine multi-task-learning and NCDE architecture to perform both anomaly detection, and PoA detection and achieve the best performance through task-specific-parameter sharing. Additionally, we propose a novel dual co-evolving NCDE structure. Two NCDEs perform anomaly detection and PoA detection tasks. Our experiments with the 3 real-world datasets and 17 baseline methods successfully prove the efficacy of the proposed concept. In addition, our visualization of anomaly detection results delivers how our proposed method works. In the future work, we plan to conduct unsupervised precursor-of-anomaly detection research since the time series data augmentation method requires a pre-processing step.

\section*{Acknowledgements}
Noseong Park is the corresponding author. 
This work was supported by the Yonsei University Research Fund of 2021, and the Institute of Information \& Communications Technology Planning \& Evaluation (IITP) grant funded by the Korean government (MSIT) (No. 2020-0-01361, Artificial Intelligence Graduate School Program (Yonsei University), and (No.2022-0-00857, Development of AI/data-based financial/economic digital twin platform,10\%) and (No.2022-0-00113, Developing a Sustainable Collaborative Multi-modal Lifelong Learning Framework, 45\%),(2022-0-01032, Development of Collective Collaboration Intelligence Framework for Internet of Autonomous Things, 45\%).
\bibliographystyle{ACM-Reference-Format}
\bibliography{sample-base}

\clearpage
\appendix
\section{Baselines}
We list all the hyperparameter settings for baselines and our method in Appendix. We compare our model with the following 16 baselines of 5 categories, including not only traditional methods but also state-of-the-art deep learning-based models as follows:
\begin{enumerate}
\item The classical method category includes the following algorithms:
\begin{enumerate}
\item OCSVM ~\citep{tax2004support} aims to find optimized support vectors that can accurately explain given data. 
\item Isolation Forest ~\citep{liu2008isolation} is basically from a decision tree. It detects anomalies based on density. 
\end{enumerate}
\item The clustering-based method category has the following methods:
\begin{enumerate}
\item Deep support vector data description (Deep-SVDD)~\citep{ruff2019deep} is a model that applies deep learning to SVDD for the anomaly detection. This method detects anomalies that are far from the compressed representation of normal data.
\item Integrative tensor-based anomaly detection (ITAD)~\citep{shin2020itad} is a tensor-based model. This model uses not only tensor decomposition but also k-means clustering to distinguish normal and abnormal. 
 \item Temporal hierarchical one-class (THOC)~\citep{shen2020timeseries} 
utilizes a dilated RNN with skip connections to capture dynamics of time series in multiple scales. Multi-resolution temporal clusters are helpful for the anomaly detection.
\end{enumerate}
\item The density-estimation-based methods are as follows:
\begin{enumerate}
\item LOF ~\citep{breunig2000lof} considers the relative density of the data and considers data with low density as abnormal data points.
\item Deep auto-encoding gaussian mixture model (DAGMM)  \citep{zong2018deep} consists of a compression network and an estimation network. Each network measures information necessary for the anomaly detection and the likelihood of the information.

\item MPPCACD~\citep{yairi2017data} is a kind of GMM and employs probabilistic dimensionality reduction and clustering to detect anomalies. 
\end{enumerate}
\item The reconstruction-based method category has the following methods:
\begin{enumerate}
    \item VAR ~\citep{Anderson1976TimeSeries2E} adapted ARIMA to anomaly detection. 
     Detect anomalies as prediction errors for future observations.
    \item LSTM ~\citep{hundman2018detecting} is a kind of RNN and is widely used for time series forecasting, because it is an algorithm for learning long-time dependencies.
    \item CL-MPPCA~\citep{tariq2019detecting} exploits convolutional LSTM to forecast future observations and mixtures of probabilistic principal component analyzers (MPPCA) to complement the convolutional LSTM.
    \item LSTM-VAE~\citep{park2018multimodal} is a model composed of an LSTM-based variational autoencoder(VAE). The decoder reconstructs the expected distribution of input. An anomaly score is measured with the estimated distribution.
    \item BeatGAN~\citep{zhou2019beatgan} trains an autoencoder with a discriminator and an adversarial loss like GANs~\citep{ganian}. Reconstruction errors are used as anomaly scores.
    \item OmniAnomaly~\citep{su2019robust} model the temporal dependency between stochastic variables by combining GRU, VAE, and planar normalizing flows. Detect anomalies based on the reconstruction probability. 
    \item Unsupervised anomaly detection (USAD)~\citep{audibert2020usad} is an auto-encoder architecture based on adversarial training. This method detects anomalies based on reconstruction errors.
    \item InterFusion~\citep{li2021multivariate} uses hierarchical VAE with two stochastic latent variables for inter-metric or temporal embeddings. Furthermore, MCMC-based method helps to obtain more reasonable reconstructions.
    \item Anomaly Transformer~\citep{xu2021anomaly} incorporates the innovative "Anomaly-Attention" mechanism to enhance the distinguishability, which computes the association discrepancy, and employs a minimax strategy.
    
\end{enumerate}
\end{enumerate}

\section{Hyperparameters for Our Model}
We test the following common hyperparameters for our method:
\begin{enumerate}
    \item In MSL, we train for 100 epochs, a learning rate of $\{1.0\times e^{-2},1.0\times e^{-3},1.0\times e^{-4}\}$, a weight decay of $\{1.0 \times e^{-3},1.0 \times e^{-4},1.0 \times e^{-5}\}$, and a size of hidden vector size is $\{19,29,39,49\}$. 
    \item In SWaT, we train for 100 epochs, a learning rate of $\{1.0\times e^{-2}, 1.0\times e^{-3},1.0\times e^{-4}\}$, a weight decay of $\{1.0\times e^{-3},1 \times e^{-4},1\times e^{-5}\}$, and a size of hidden vector size is $\{39,49,59\}$. 
    \item In WADI, we train for 100 epochs, a learning rate of $\{1.0\times e^{-2},1.0\times e^{-3},1.0\times e^{-4}\}$, a weight decay of $\{1.0 \times e^{-3},1.0 \times e^{-4},1.0 \times e^{-5}\}$, and a size of hidden vector size is $\{29,39,49,59\}$.
\end{enumerate}
In Table~\ref{tbl:msl_archi1} to ~\ref{tbl:wadi_archi3}, we clarify the network architecture of the CDE functions, $f$,$g$, and $c$. 

\begin{table}[t]
\setlength{\tabcolsep}{4pt}
\caption{The best architecture of the CDE function $f$ for MSL. $\texttt{FC}$, $\phi$,$\rho$ stands for the fully-connected layer, the rectified linear unit (ReLU), and the hyperbolic tangent (tanh), respectively.}\label{tbl:msl_archi1}
\begin{tabular}{cccc}
\hline
Design                  & Layer & Input                 & Output            \\ \hline
$\phi$(\texttt{FC})     & 1     & $256  \times 64 $   & $256 \times 256  $\\
$\phi$(\texttt{FC})     & 2     & $256  \times 256 $   & $256 \times 256  $\\
$\phi$(\texttt{FC})     & 3     & $256  \times 256 $   & $256 \times 256  $\\
$\phi$(\texttt{FC})     & 4     & $256  \times 256 $   & $256 \times 256  $\\
\texttt{FC}     & 5     & $256  \times 256$   & $256 \times 3,520$  \\\hline
\end{tabular}
\end{table}

\begin{table}[t]
\setlength{\tabcolsep}{4pt}
\caption{The best architecture of the CDE function $g$ for MSL}\label{tbl:msl_archi2}
\begin{tabular}{cccc}
\hline
Design                  & Layer & Input                 & Output            \\ \hline
$\phi$(\texttt{FC})            & 1     & $256  \times 64 $   & $256 \times 512   $\\
$\phi$(\texttt{FC})     & 2     & $256  \times 512 $   & $256 \times 512  $\\
$\phi$(\texttt{FC})     & 3     & $256  \times 512 $   & $256 \times 512  $\\
$\phi$(\texttt{FC})     & 4     & $256  \times 512 $   & $256 \times 512  $\\
\texttt{FC}     & 5     & $256  \times 512$   & $256 \times 3,520$  \\\hline
\end{tabular}
\end{table}

\begin{table}[t]
\setlength{\tabcolsep}{4pt}
\caption{The best architecture of the shared CDE function $c$ for MSL}\label{tbl:msl_archi3}
\begin{tabular}{cccc}
\hline
Design                  & Layer & Input                 & Output            \\ \hline
$\phi$(\texttt{FC})     & 1     & $256  \times 64 $   & $256 \times 64  $\\
$\rho$(\texttt{FC})     & 2     & $256  \times 256$   & $256 \times 3,520$  \\\hline
\end{tabular}
\end{table}

\begin{table}[t]
\setlength{\tabcolsep}{4pt}
\caption{The best architecture of the CDE function $f$ for SWaT.}\label{tbl:swat_archi}
\begin{tabular}{cccc}
\hline
Design                  & Layer & Input                 & Output            \\ \hline
$\phi$(\texttt{FC})     & 2     & $256  \times 64 $   & $256 \times 128  $\\
$\phi$(\texttt{FC})     & 3     & $256  \times 128 $   & $256 \times 128  $\\
$\phi$(\texttt{FC})     & 4     & $256  \times 128 $   & $256 \times 128  $\\
$\phi$(\texttt{FC})     & 5     & $256  \times 128 $   & $256 \times 128  $\\
\texttt{FC}     & 6     & $256  \times 128$   & $256 \times 3,264$  \\\hline
\end{tabular}
\end{table}

\begin{table}[t]
\setlength{\tabcolsep}{4pt}
\caption{The best architecture of the CDE function $g$ for SWaT.}\label{tbl:swat_archi2}
\begin{tabular}{cccc}
\hline
Design                  & Layer & Input                 & Output            \\ \hline
$\phi$(\texttt{FC})     & 2     & $256  \times 64 $   & $256 \times 64  $\\
$\phi$(\texttt{FC})     & 3     & $256  \times 64 $   & $256 \times 64  $\\
$\phi$(\texttt{FC})     & 4     & $256  \times 64 $   & $256 \times 64  $\\
$\phi$(\texttt{FC})     & 5     & $256  \times 64 $   & $256 \times 64  $\\
\texttt{FC}     & 6     & $256  \times 64$   & $256 \times 3,264$  \\\hline
\end{tabular}
\end{table}

\begin{table}[t]
\setlength{\tabcolsep}{4pt}
\caption{The best architecture of the shared CDE function $c$ for SWaT.}\label{tbl:swat_archi3}
\begin{tabular}{cccc}
\hline
Design                  & Layer & Input                 & Output            \\ \hline
$\phi$(\texttt{FC})     & 1     & $256  \times 64 $   & $256 \times 64  $\\
$\rho$(\texttt{FC})     & 2     & $256  \times 64$   & $256 \times 3,264$  \\\hline
\end{tabular}
\end{table}

\begin{table}[t]
\setlength{\tabcolsep}{4pt}
\caption{The best architecture of the CDE function $f$ for WADI.}\label{tbl:wadi_archi}
\begin{tabular}{cccc}
\hline
Design                  & Layer & Input                 & Output            \\ \hline
$\phi$(\texttt{FC})     & 1     & $256  \times 16 $   & $256 \times 128  $\\
$\phi$(\texttt{FC})     & 2     & $256  \times 128 $   & $256 \times 128  $\\
$\phi$(\texttt{FC})     & 3     & $256  \times 128 $   & $256 \times 128  $\\
$\phi$(\texttt{FC})     & 4     & $256  \times 128 $   & $256 \times 128  $\\
\texttt{FC}    & 5     & $256  \times 128$   & $256 \times 1,968 $  \\\hline
\end{tabular}
\end{table}

\begin{table}[t]
\setlength{\tabcolsep}{4pt}
\caption{The best architecture of the CDE function $g$ for WADI.}\label{tbl:wadi_archi2}
\begin{tabular}{cccc}
\hline
Design                  & Layer & Input                 & Output            \\ \hline
$\phi$(\texttt{FC})     & 1     & $256  \times 16 $   & $256 \times 128  $\\
$\phi$(\texttt{FC})     & 2     & $256  \times 128 $   & $256 \times 128  $\\
$\phi$(\texttt{FC})     & 3     & $256  \times 128 $   & $256 \times 128  $\\
$\phi$(\texttt{FC})     & 4     & $256  \times 128 $   & $256 \times 128  $\\
\texttt{FC}     & 5     & $256  \times 128 $   & $256 \times 1,968$  \\\hline
\end{tabular}
\end{table}

\begin{table}[t]
\setlength{\tabcolsep}{4pt}
\caption{The best architecture of the shared CDE function $c$ for WADI.}\label{tbl:wadi_archi3}
\begin{tabular}{cccc}
\hline
Design                  & Layer & Input                 & Output            \\ \hline
$\phi$(\texttt{FC})     & 1     & $256  \times 16 $   & $256 \times 256  $\\
$\rho$(\texttt{FC})     & 2     & $256  \times 256$   & $256 \times 1,968$  \\\hline
\end{tabular}
\end{table}

\section {Hyperparameters for baselines}
For the best outcome of baselines, we conduct hyperparmeter search for them based on the recommended hyperparameter set from each papers. 
\begin{enumerate}
    \item Classical methods: For OCSVM, we use the RBF kernel. For Isolation Forest, we use the base estimator of $100$ in the ensemble, and choose the number of trees in $\{25,100\}$.
    \item Clustering-based methods: For Deep SVDD and ITAD, we use a learning rate of $\{1.0 \times e^{-3},1.0 \times e^{-4}\}$ and a hidden vector dimension of $\{32,64,128\}$ and For THOC, we use a number of hidden units in $\{32,64,84\}$, and we use its default hyperparameters. 
    
    \item Density-estimation methods: For LOF, we use the number of neighbors in $\{1,3,5,7,12\}$. For DAGMM and MMPCACD, we follow those default hyperparameters.
    
    \item Reconstruction-based methods: For LSTM, we use a learning rate of $\{1.0 \times e^{-3},1.0 \times e^{-4}\}$ and a hidden vector dimension of $\{80,100,120\}$. For LSTM-VAE, BeatGAN, and OmniAnomaly, we use a learning rate of $\{1.0 \times e^{-4},1.0 \times e^{-5}\}$ and a hidden vector dimension of $\{32,64,128\}$. For USAD, we use a learning rate of $\{1.0 \times e^{-3},1.0 \times e^{-4}\}$ and a hidden vector dimension of $\{32,64,128\}$, and follow other default hyperparameters in USAD. For InterFusion, a hidden vector dimension of $\{128,256,512\}$, and follow other default hyperparameters in InterFusion. For Anomaly Transformer, we follow all hyperparameters in Anomaly Transformer.
    
\end{enumerate}

\section{Data preprocessing details}
We use the following datasets to compare PAD with other methods: 
\begin{enumerate}
    \item Mars Science Laboratory: MSL is a public dataset from NASA. MSL is licensed under the following license:  \url{https://github.com/khundman/telemanom/blob/master/LICENSE.txt}
    \item Secure Water Treatment: SWaT is licensed under the following license: \url{https://itrust.sutd.edu.sg/testbeds/secure-water-treatment-swat/}
    \item Water Distribution: WADI is a public dataset from NASA. WADI is licensed under the following license:  \url{https://itrust.sutd.edu.sg/testbeds/water-distribution-wadi/}
\end{enumerate}

Since our method resorts to a self-supervised multi-task learning approach, we augmented training samples with abnormal patterns and its detailed process is in Alg.~\ref{alg:dataaug}. The augmentation method is similar to other popular augmentation methods for images, e.g., CutMix~\cite{yun2019cutmix}. There is a long training sequence $\mathbf{x}_{0:T}=\{\mathbf{x}_0,\mathbf{x}_1,...,\mathbf{x}_T\}$. We apply the augmentation method to the raw sequence $\mathbf{x}_{0:T}$ before segmenting it into windows. We randomly copy existing $l$ observations from a random location $r$ to a target location $s$. In general, the ground-truth anomaly pattern is unknown in each dataset. Although our copy-and-paste augmentation method is simple, our experimental results prove its effectiveness. At the same time, we also believe that there will be better augmentation methods.

\begin{algorithm}[t]
\SetAlgoLined
\caption{How to apply our augmentation method to training samples}\label{alg:dataaug}
\KwIn{Training sequence $\mathbf{x}_{0:T}=\{\mathbf{x}_0,\mathbf{x}_1,...,\mathbf{x}_T\}$, Anomaly ratio $\gamma$}

\tcc{Init. the length after the augmentation}
$L \gets T$\;
\While {$\frac{L-T}{T} \leq \gamma$}{
    Randomly select a starting point $s$;\\
    Randomly select a resampling point $r$;\\
    Randomly select a length $l$ between $100$ and $500$\\
    \tcc{Implant abnormal observations}
    Concatenate $\mathbf{x}_{0:s}$, $\mathbf{x}_{r:r+l}$, and $\mathbf{x}_{s_i+1:N}$;\\
    \tcc{Increase the length}
    $L \gets L + l$
    }
\end{algorithm}

\end{document}